\pdfoutput=1

\documentclass[11pt]{article}

\usepackage[final]{acl}
\usepackage{booktabs}

\usepackage{times}
\usepackage{latexsym}
\usepackage{amsmath}
\usepackage[utf8]{inputenc} 
\usepackage[T1]{fontenc}    
\usepackage{hyperref}       
\usepackage{url}            
\usepackage{booktabs}       
\usepackage{amsfonts}       
\usepackage{nicefrac}       
\usepackage{microtype}      
\usepackage{xcolor}         
\usepackage{amssymb}
\usepackage{comment}
\usepackage{multirow}
\usepackage{array}
\usepackage{graphicx}
\usepackage{subcaption}
\usepackage[T1]{fontenc}

\usepackage[utf8]{inputenc}

\usepackage{booktabs}

\usepackage{microtype}

\usepackage{inconsolata}

\usepackage{graphicx}

\title{Efficient Knowledge Editing via Minimal Precomputation}

\author{$\text{Akshat Gupta}^1$, $\text{Maochuan Lu}^1$ $\text{Thomas Hartvigsen}^2$, $\text{Gopala Anumanchipalli}^1$\\
$^1\text{UC Berkeley}$, $^2\text{University of Virginia}$\\
  \texttt{akshat.gupta@berkeley.edu}}

\begin{document}
\maketitle
\begin{abstract}
Knowledge editing methods like MEMIT are able to make data and compute efficient updates of factual knowledge by using a single sentence to update facts and their consequences. However, what is often overlooked is a ``precomputation step'', which requires a one-time but significant computational cost. The authors of MEMIT \cite{MEMIT} originally precompute approximately 44 million hidden vectors per edited layer, which requires a forward pass over 44 million tokens. For GPT-J (6B) \cite{gpt-j}, this precomputation step takes 36 hours on a single GPU, while it takes approximately 40 hours for Llama2-7B \cite{llama2}. Additionally, this precomputation time grows with model size. In this paper, we show that this excessive computational cost is unnecessary. Knowledge editing using MEMIT and related methods, such as ROME and EMMET \cite{rome, akshat-unified}, can be performed by pre-computing a very small portion of the 44 million hidden vectors. We first present the theoretical minimum number of hidden vector precomputation required for solutions of these editing methods to exist. We then empirically show that knowledge editing using these methods can be done by pre-computing significantly fewer hidden vectors. Specifically, we show that the precomputation step can be done with less than 0.3\% of the originally stipulated number of hidden vectors. This saves a significant amount of precomputation time and allows users to begin editing new models within a few minutes.
\end{abstract}

\section{Introduction}
Knowledge editing \cite{editing-survey}, or the ability to edit knowledge stored within the parameters of large language models (LLMs), is a topic of growing interest. A specific type of parameter-modifying knowledge editing methods called "locate-then-edit" methods \cite{editing-survey} allow us to edit any transformer-based LLMs without the need for additional training. The most popular of these methods are MEMIT \cite{MEMIT}, the first successful method that allows for batched editing, its predecessor ROME \citep{rome, akshat-rebuilding}, which allows only one edit at a time, and EMMET \cite{akshat-unified}, that generalizes ROME to batched editing. 

While these "locate-then-edit" methods do not require additional training, we cannot just start editing a newly launched LLM instantly \cite{akshat-llama3}. Each of MEMIT, ROME, and EMMET has a precomputation step where a large number of Wikipedia articles are passed through the model being edited and the intermediate hidden representations of the edited layers are cached. The editing loss function aims to preserve the outputs of these cached hidden representations during the editing process \cite{akshat-unified}. Although this needs to be done only once to edit a model, it still requires a significant computational overhead. For example, the original authors of MEMIT pre-computed about 44 million hidden vectors per edited layer. This computation takes about 36 hours for GPT-J (6B) \cite{gpt-j} and 40 hours for Llama-2 (7B) \cite{llama2} on a single GPU\footnote{Numbers calculated on a single NVIDIA A6000 GPU with 48 GB GPU memory.}. Additionally, these numbers increase with the size of the model and the number of layers being edited. This means that while locate-then-edit methods do not require additional training and can be very quick during inference, they do require a significant initial computational cost which grows with the model size.

In this paper, we show that this large amount of precomputation is unnecessary. We first analyze the closed-form solution for the different editing algorithms and find the theoretical minimum amount of tokens required  for the precomputation step. We then empirically show that optimal knowledge editing performance can be achieved by performing precomputation on approximately twice this minimum number. This allows us to achieve comparable knowledge editing performance using less than 0.1\% of the originally stipulated 44 million tokens for GPT2-XL and GPT-J. We call the efficient versions of these methods as the FastMEMIT family of editing methods, which significantly reduce the upfront computation costs, making it possible to begin editing models within minutes. We also release our code, which can be found here - \url{https://github.com/scalable-model-editing/efficient-model-editing}.



\section{Background}
In ``locate-then-edit`` knowledge editing methods, facts for model editing are usually represented in a key-value format, where the key vector helps locate a fact, and the value vector provides the target output after editing \citep{MEMIT}. For example, for the edited fact "The capital of Malaysia is Singapore," $k_e$ corresponds to the query "The capital of Malaysia is," and $v_e$ corresponds to the new target ``Singapore.'' Additionally, $k_0$ represents key vectors whose outputs need to remain constant during editing, ensuring the editing process doesn't impact the general ability \citep{akshat-catastrophic} or unrelated knowledge of edited models. 

During editing, we first identify the layer that is maximally responsible for retrieving a fact, and then update the corresponding weight matrix to reflect the updated fact. In this process, we want to make sure two things: one is to preserve previously stored knowledge, and the other one is to memorize what we edit. For MEMIT \citep{MEMIT}, causal mediation analysis showed that the MLP modules within certain layers are responsible for storing factual knowledge. The knowledge editing objective of MEMIT is formulated as follows \cite{akshat-unified}: 

\vskip -0.2in
\begin{equation}\label{eq:memit_objective}
     \underset{\hat{W}}{\operatorname{argmin}} \hspace{4pt} \underbrace{\lambda \left\| \hat{W} K_0 - W_0 K_0 \right\|^2_F}_{\text{preservation}}  + \underbrace{\left\|\hat{W} K_E - V_E \right\|^2_F}_{\text{memorization}}
\end{equation}

The above loss can be interpreted as a summation of two terms. In the first term, we preserve the outputs for a collection of input key-vectors ($K_0$) to preserve the existing knowledge of the model, while in the second term we force the outputs of certain key-vectors ($K_E$) to a target ($V_E$). The argument $\hat{W}$ is the second MLP matrix in the FFN module of a transformer. Since the above objective is linear in the argument, we can derive a closed form solution, as shown below:

\vskip -0.1in
\begin{equation}\label{eq:memit}
\begin{aligned}
    \hat{W} &= W_0 + \Delta \hspace{10pt} \text{where} \hspace{10pt}  
    \\ \Delta &= \big(V_E - W_0K_E \big) K_E^T \big( \lambda C_0 + K_EK_E^T \big)^{-1}
\end{aligned}
\end{equation}

where $W_0$ is the unedited weight matrix, and $\hat{W}$ refers to the updated weights. $k_0$ denotes the key-vector for preserving knowledge from the original model. $K_0 = [k^0_1 \hspace{4pt}| k^0_2 \hspace{4pt}| \dots |\hspace{4pt}k^0_P]$ is a matrix containing all these preserved key-vectors. $k_e$ denotes the key-vectors representing modified facts, and $K_E = [k^e_1 \hspace{4pt}| k^e_2 \hspace{4pt}| \dots |\hspace{4pt}k^e_B]$ is a matrix containing edited key-vectors. The output at the edited layer corresponding to $k_e$ is denoted by $v_e$ and $V_E = [v^e_1 \hspace{4pt}| v^e_2 \hspace{4pt}| \dots |\hspace{4pt}v^e_B]$ is the matrix containing all  target vectors.

\subsection{Overview of knoweldge editing metrics}\label{sec:metrics}
In this paper, we evaluate knowledge editing methods using the following standard knowledge editing metrics \cite{MEMIT}:

\begin{itemize}
    \item \textbf{Efficacy Score (ES)} evaluates the success of an edit. It is calculated as the percentage of edits for which $P(\text{new fact}) > P(\text{old fact})$.
    
    \item \textbf{Paraphrase Score (PS)} evaluates the model’s generalization ability for an edit, calculated as the $P(\text{new fact}) > P(\text{old fact})$ when a paraphrase of the editing prompt is used as query.
    \item \textbf{Neighborhood Score (NS)} evaluates the locality or specificity of an edit. It is calculated as the percentage of the facts in the neighborhood of the edited fact that remain unchanged after an edit.
    \item \textbf{Overall Score (S)} is the harmonic mean of ES, PS, and NS. 
\end{itemize}

\section{Dataset and Models}
We perform singular and batch editing experiments on the CounterFact dataset \cite{rome}. CounterFact is a standard dataset used in knowledge editing. We peform knowledge editing on three representative models - GPT2-XL \cite{gpt-2}, GPT-J (6B) \cite{gpt-j} and Llama2-7B \cite{llama2}.

\section{Theoretical Minimum Tokens for Precomputation}
One major benefit of the closed-form solution in MEMIT is the presence of the covariance matrix, $C_0 = K_0K_0^T$, which can be written as a sum of outer products of key-vectors as shown below: 

\vskip -0.2 in
\begin{equation}\label{eq:covariance_matrix}
    C_0 = K_0 K^T_0 = \sum^P_{i=1} k^i_0 k^{i^T}_0
\end{equation}

Here, $P$ denotes the number of preserved vectors in equation \ref{eq:memit_objective}. This matrix $C_0$ remains fixed during editing since it is made up of key-vectors that serve as the input of the edited matrix, which is why $C_0$ is precomputed before editing begins. $C_0$ is one part of the matrix that gets inverted in the closed-form solution of MEMIT (equation \ref{eq:memit}). If we represent the matrix being inverted in the closed form solution as $C_{\text{eff}}$, then:
\begin{equation}\label{eq:covariance_eff_matrix}
\begin{aligned}
    C_{\text{eff}} &= \lambda K_0 K^T_0 + K_E K^T_E \\
    &= \lambda \sum^P_{i=1} k^i_0 k^{i^T}_0 + \sum^B_{i=1} k^i_e k^{i^T}_e
\end{aligned}
\end{equation}

A pre-requisite of the closed-form solution to exist is the invertibility of the $C_{\text{eff}}$ matrix. As shown above, $C_{\text{eff}}$ matrix is a sum of outer products of $P+B$ vectors, where $B$ represents the batch size for editing. For a model with hidden dimension $d$, the dimensionality of a key-vector is usually $4d$. This means that the $C_{\text{eff}}$ matrix is a square matrix of dimensionality $4d$. For a $4d$-dimensional square matrix which is a summation of rank-1 matrices, it is invertible as long as there are at least $4d$-independent vectors in the summation. For example, for GPT2-XL with hidden dimension of 1600, the dimensionality of key vectors are 6400. Thus, as long as representations of at least 6400 independent key-vectors are preserved or memorized while editing, $C_{\text{eff}}$ will be an invertible matrix. \textit{This is a fundamental assumption in MEMIT.}

We want to find the minimum number of keys that need to be preserved in order for $C_{\text{eff}}$ to be invertible.
Since the editing batch size ($B$) is varied from one to larger batch sizes, we take $B=1$ for this argument. If we let the dimensionatity of the key-vectors be $d_k$, then for an editing batch size of 1, \textbf{at least $d_k-1$ key-vectors need to be preserved, granted they are \underline{independent} of each other}. This number serves as the theoretical minimum number of tokens over which we need to perform precomputations.

In practice, MEMIT preserves the representations of a much larger number of vectors - 44 million tokens to be specific. For each layer being edited, this step takes about 1.5 hours for GPT-XL, 6 hours for GPT-J, 8 hours for Llama-2-7B. Since multiple layers are edited within a model in MEMIT, this number usually requires tens of hours of precomputation, and scales linearly with the size of the model being edited\footnote{Numbers calculated for 1 RTX A6000 48GB GPU}.

\begin{figure*}[h!]
    \centering
    \begin{subfigure}{0.24\linewidth}
        \centering
        \includegraphics[width=\linewidth]{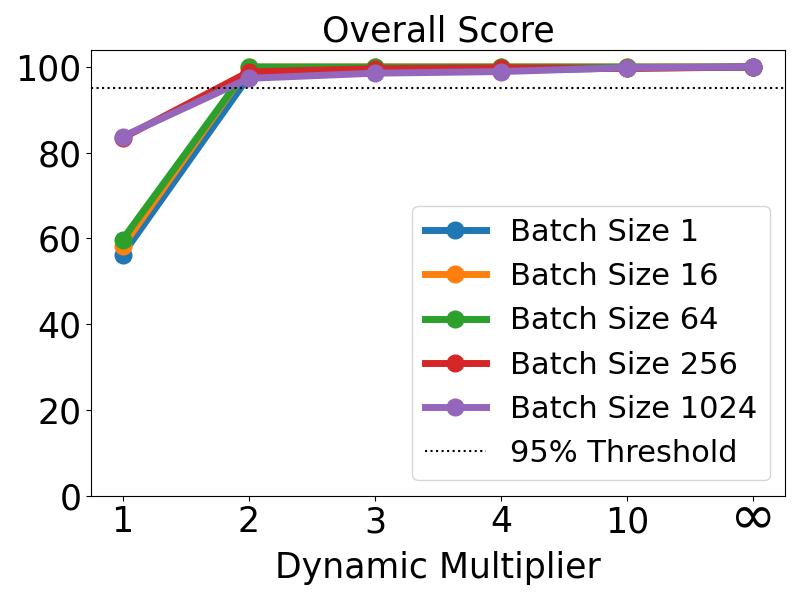}
        \caption{Overall Score}
        \label{fig:FastEMMET-gptj-overall}
    \end{subfigure}
    \hfill
    \begin{subfigure}{0.24\linewidth}
        \centering
        \includegraphics[width=\linewidth]{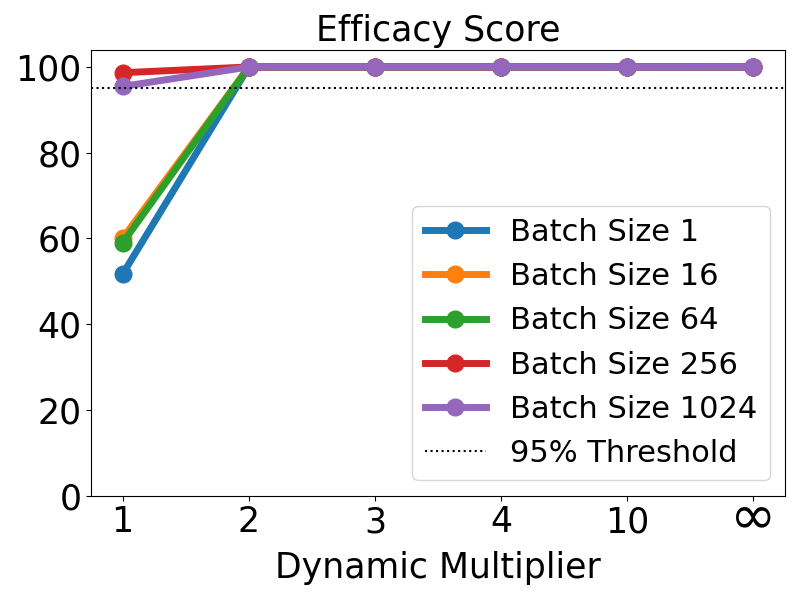}
        \caption{Efficacy score}
        \label{fig:FastEMMET-gptj-efficacy}
    \end{subfigure}
    \hfill
    \begin{subfigure}{0.24\linewidth}
        \centering
        \includegraphics[width=\linewidth]{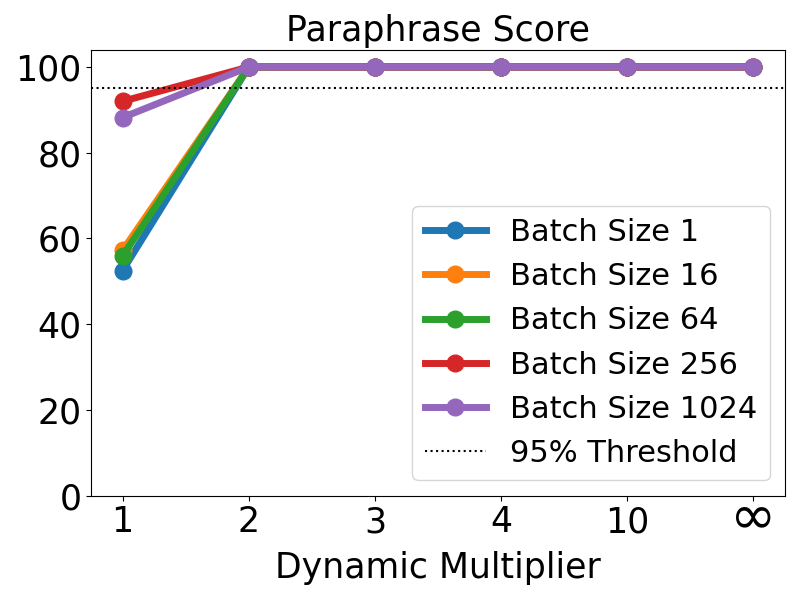}
        \caption{Paraphrase score}
        \label{fig:FastEMMET-gptj-paraphrase}
    \end{subfigure}
    \hfill
    \begin{subfigure}{0.24\linewidth}
        \centering
        \includegraphics[width=\linewidth]{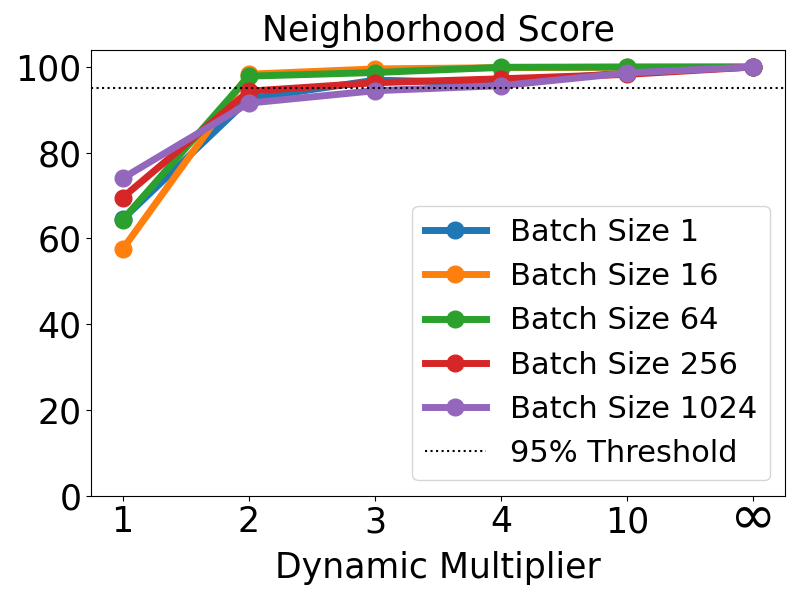}
        \caption{Neighborhood score}
        \label{fig:FastEMMET-gptj-neighborhood}
    \end{subfigure}
    \caption{Performance of FastEMMET in GPT-J across different batch sizes}
    \label{fig:FastEMMET-performance-gptj}
\end{figure*}

\begin{figure*}[h!]
    \centering
    \begin{subfigure}{0.24\linewidth}
        \centering
        \includegraphics[width=\linewidth]{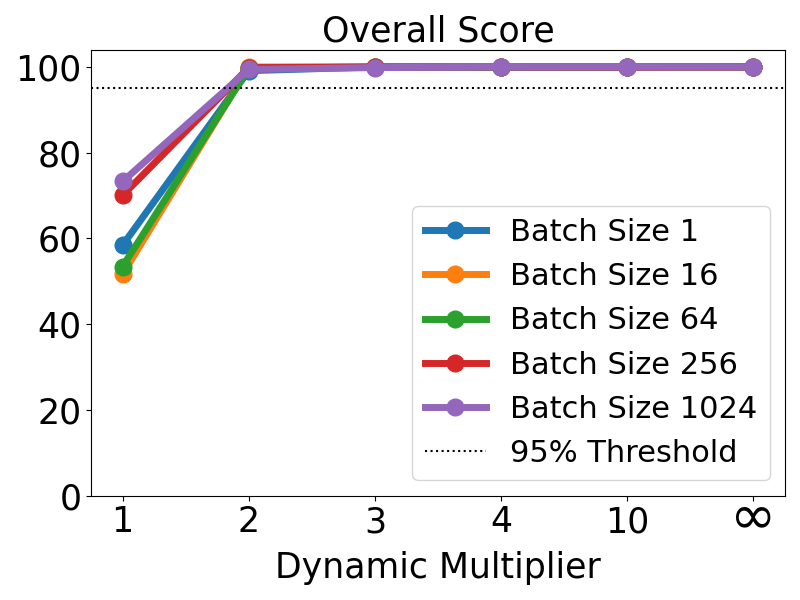}
        \caption{Overall Score}
        \label{fig:FastMEMIT-gptj-overall}
    \end{subfigure}
    \hfill
    \begin{subfigure}{0.24\linewidth}
        \centering
        \includegraphics[width=\linewidth]{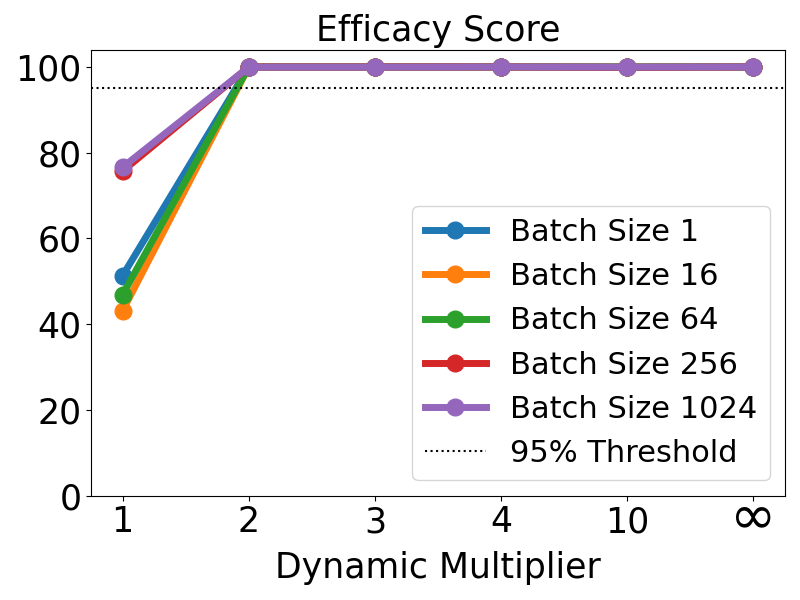}
        \caption{Efficacy score}
        \label{fig:FastMEMIT-gptj-efficacy}
    \end{subfigure}
    \hfill
    \begin{subfigure}{0.24\linewidth}
        \centering
        \includegraphics[width=\linewidth]{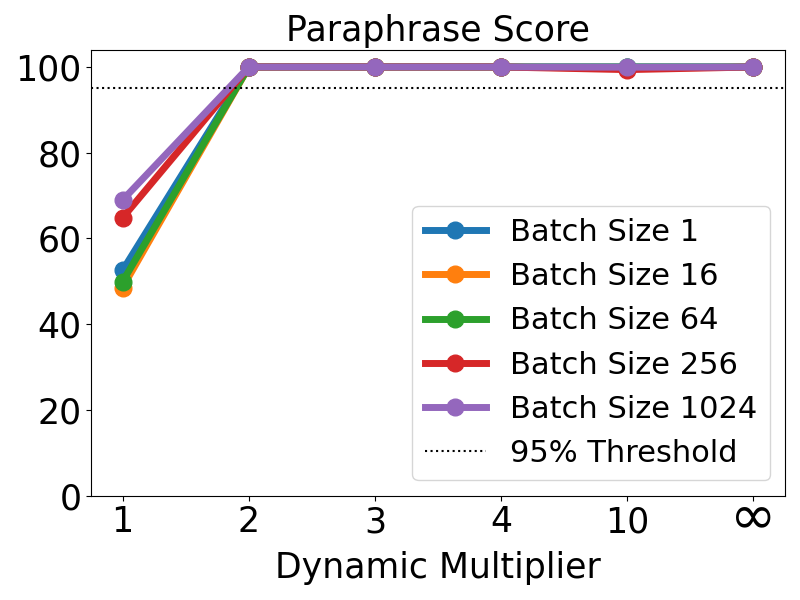}
        \caption{Paraphrase score}
        \label{fig:FastMEMIT-gptj-paraphrase}
    \end{subfigure}
    \hfill
    \begin{subfigure}{0.24\linewidth}
        \centering
        \includegraphics[width=\linewidth]{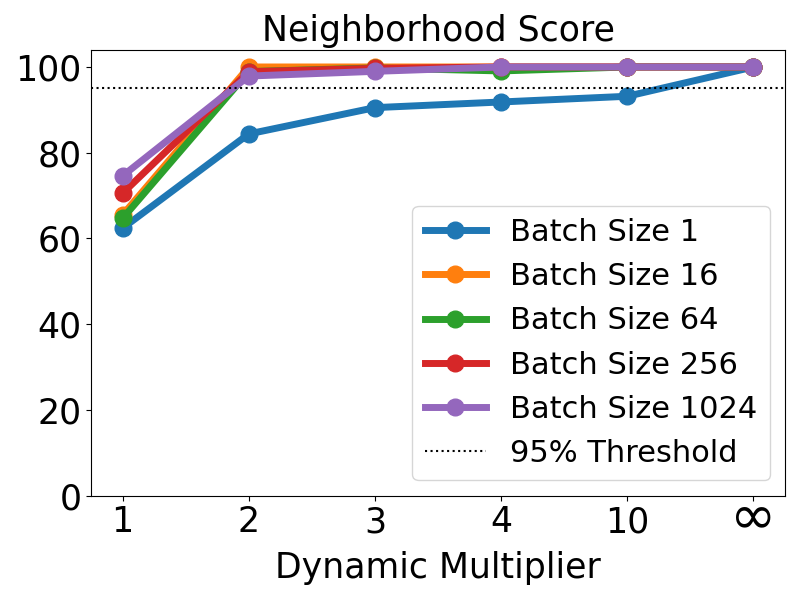}
        \caption{Neighborhood score}
        \label{fig:FastMEMIT-gptj-neighborhood}
    \end{subfigure}
    \caption{Performance of FastMEMIT in GPT-J across different batch sizes}
    \label{fig:FastMEMIT-performance-gptj}
\end{figure*}

\section{FastMEMIT Family of Methods}
In the above section, we show that the minimum number of tokens required for the computation of $C_0$ matrix is $d_k - 1$, where $d_k$ is the dimensionality of the key-vectors in an MLP. For GPT2-XL, $d_k = 6400$, whereas for GPT-J, $d_k = 16384$. While the theoretical minimum number of precomputations required is approximately equal to $d_k$, we ask the question - \textbf{"what is the optimal number of tokens required for precomputation without compromising on editing performance?"}.

We begin by using the theoretical minimum number for precomputation and quickly find that this leads to loss of editing performance. We also find that for some cases, especially for Llama2-7B models, using this theoretical minimum leads to un-invertible matrices, since the selected vectors may not be independent. We increase the number of precomputation tokens in increments of the theoretical minimum. For this, we introduce \textbf{dynamic multiplier}, a hyperparameter that controls the number of preserved key vectors in $C_0$. For example, with a dynamic multiplier of $d_m = 3$, the number of pre-computed key vectors is reduced to $3 \times d_k$, where $d_k$ is approximately equal to the theoretical minimum. This is a significantly lower computational cost while ensuring the matrix remains invertible. For example, with $d_m=3$ for GPT2-XL, the precomputation is done over 12,288 tokens, which is approximately 0.02\% of the original 44 million tokens.

With this dynamic multiplier, we can rewrite Equation~\ref{eq:covariance_eff_matrix} as follows:

\begin{equation}\label{eq:covariance_eff_matrix_new}
\begin{aligned}
    C_{\text{eff}} &= \lambda K_0 K^T_0 + K_E K^T_E \\
    &= \lambda \sum_{i=1}^{P'} k^i_0 k^{i^T}_0 + \sum_{i=1}^B k^i_e k^{i^T}_e,
\end{aligned}
\end{equation}

where $P' = d_m \cdot d_k$. The same idea of using the dynamic multiplier to reduce the number of preserved keys can be applied to ROME \cite{rome} and its batch generalization EMMET \cite{akshat-unified}. We refer to the reduced precomputation version of these methods as FastMEMIT, FastROME, and FastEMMET in this paper.



To evaluate these methods, we perform batched knowledge editing for varying batch sizes, growing from 1 to 1024. For each batch size, we take samples of multiple batches (Table \ref{table:batch-size-stats} in appendix). For example, for batch size 16, the results are calculated by averaging editing results of 10 batches. Since EMMET is a batch-editing generalization of ROME, we present the results for EMMET in this paper. The editing results for ROME correspond to EMMET with batch size 1. 

\begin{figure*}[h!]
    \centering
    \begin{subfigure}{0.24\linewidth}
        \centering
        \includegraphics[width=\linewidth]{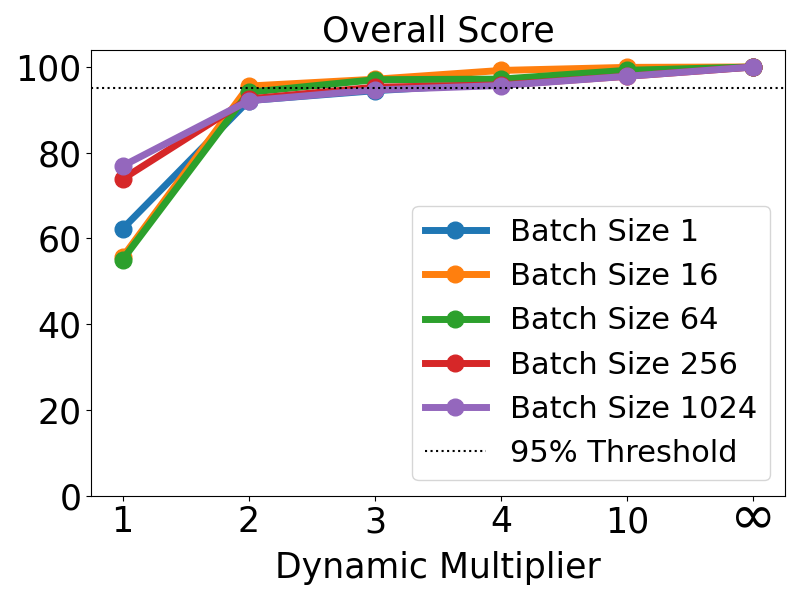}
        \caption{Overall Score}
        \label{fig:FastEMMET-llama2-overall}
    \end{subfigure}
    \hfill
    \begin{subfigure}{0.24\linewidth}
        \centering
        \includegraphics[width=\linewidth]{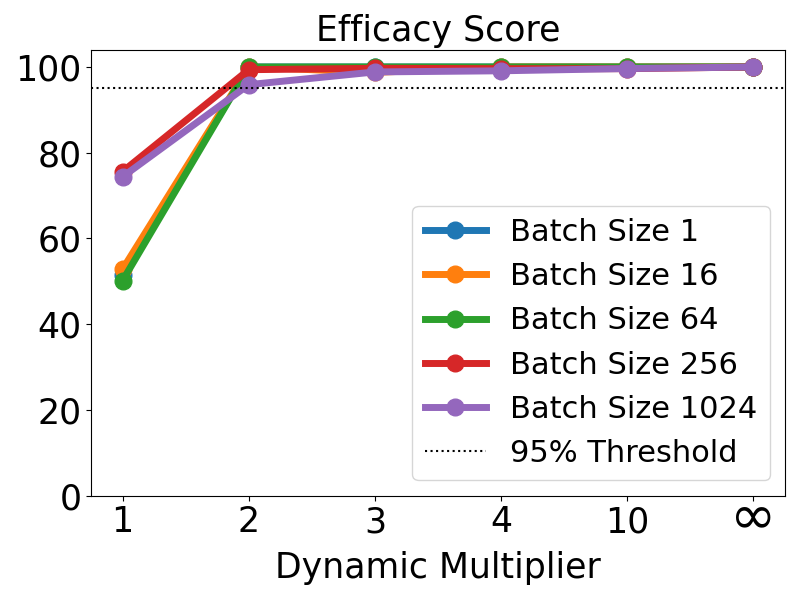}
        \caption{Efficacy score}
        \label{fig:FastEMMET-llama2-efficacy}
    \end{subfigure}
    \hfill
    \begin{subfigure}{0.24\linewidth}
        \centering
        \includegraphics[width=\linewidth]{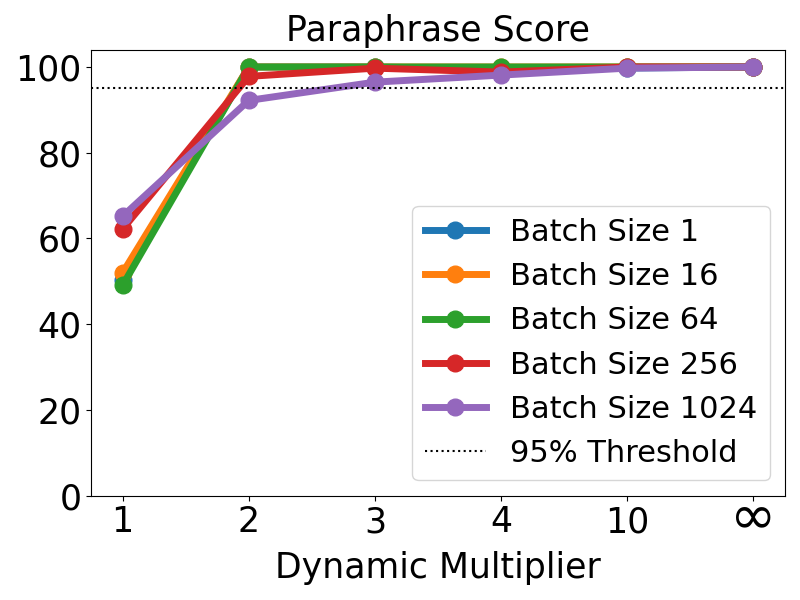}
        \caption{Paraphrase score}
        \label{fig:FastEMMET-llama2-paraphrase}
    \end{subfigure}
    \hfill
    \begin{subfigure}{0.24\linewidth}
        \centering
        \includegraphics[width=\linewidth]{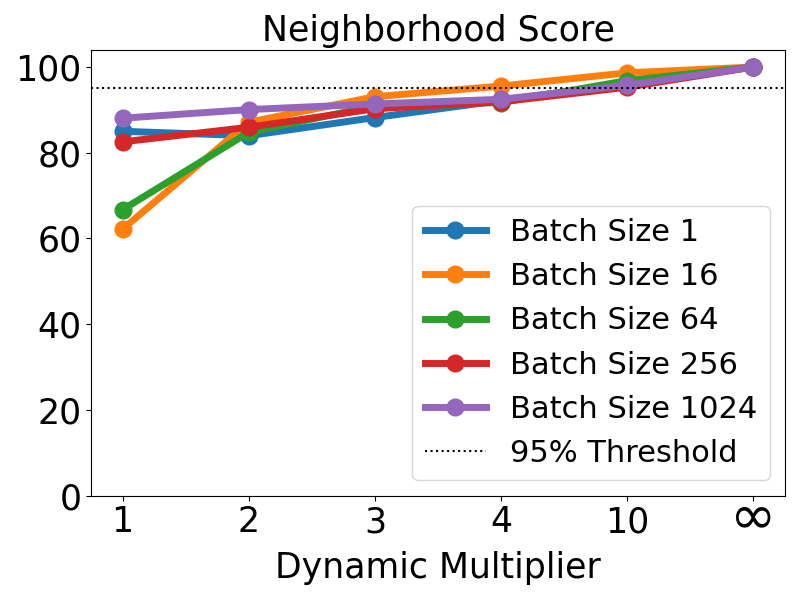}
        \caption{Neighborhood score}
        \label{fig:FastEMMET-llama2-neighborhood}
    \end{subfigure}
    \caption{Performance of FastEMMET in Llama 2 across different batch sizes}
    \label{fig:FastEMMET-performance-llama2}
\end{figure*}

\begin{figure*}[h!]
    \centering
    \begin{subfigure}{0.24\linewidth}
        \centering
        \includegraphics[width=\linewidth]{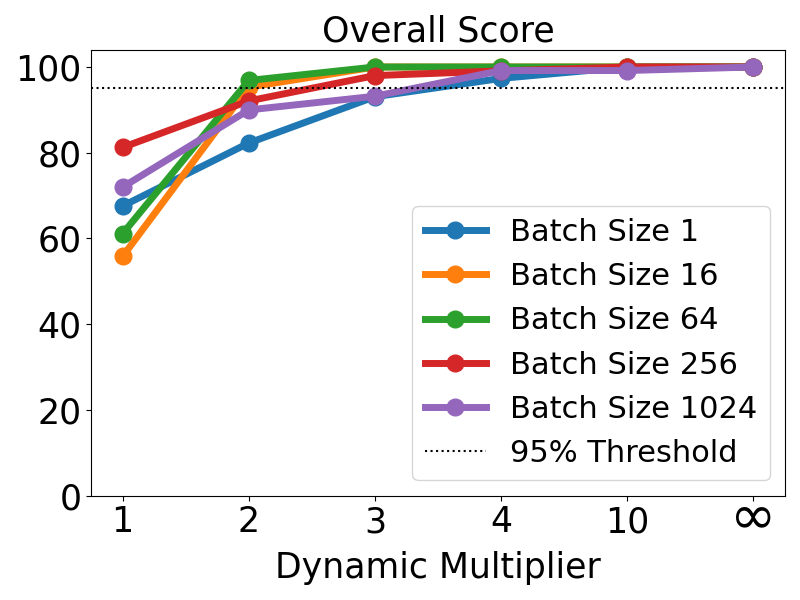}
        \caption{Overall Score}
        \label{fig:FastMEMIT-llama2-overall}
    \end{subfigure}
    \hfill
    \begin{subfigure}{0.24\linewidth}
        \centering
        \includegraphics[width=\linewidth]{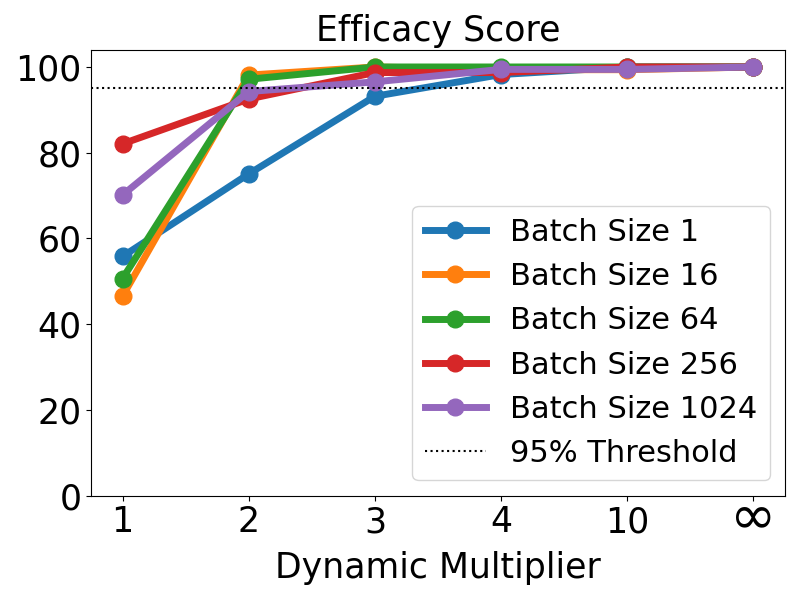}
        \caption{Efficacy score}
        \label{fig:FastMEMIT-llama2-efficacy}
    \end{subfigure}
    \hfill
    \begin{subfigure}{0.24\linewidth}
        \centering
        \includegraphics[width=\linewidth]{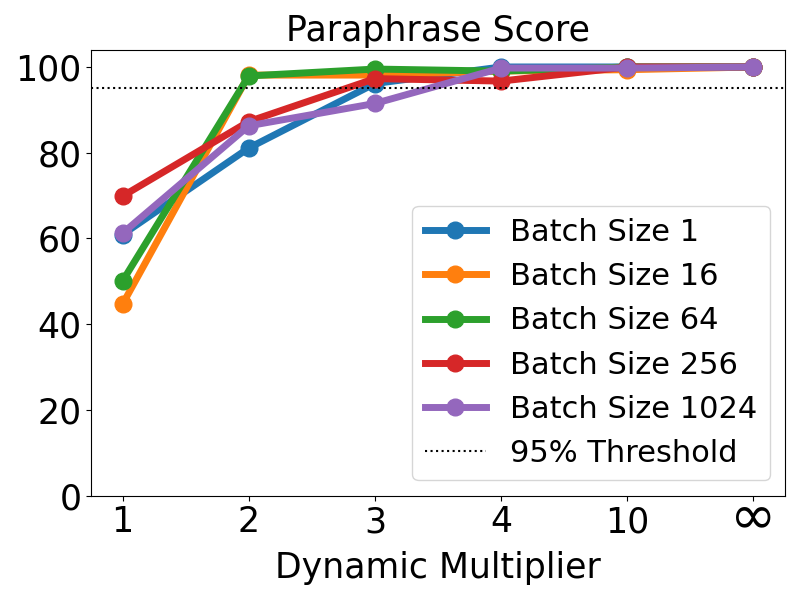}
        \caption{Paraphrase score}
        \label{fig:FastMEMIT-llama2-paraphrase}
    \end{subfigure}
    \hfill
    \begin{subfigure}{0.24\linewidth}
        \centering
        \includegraphics[width=\linewidth]{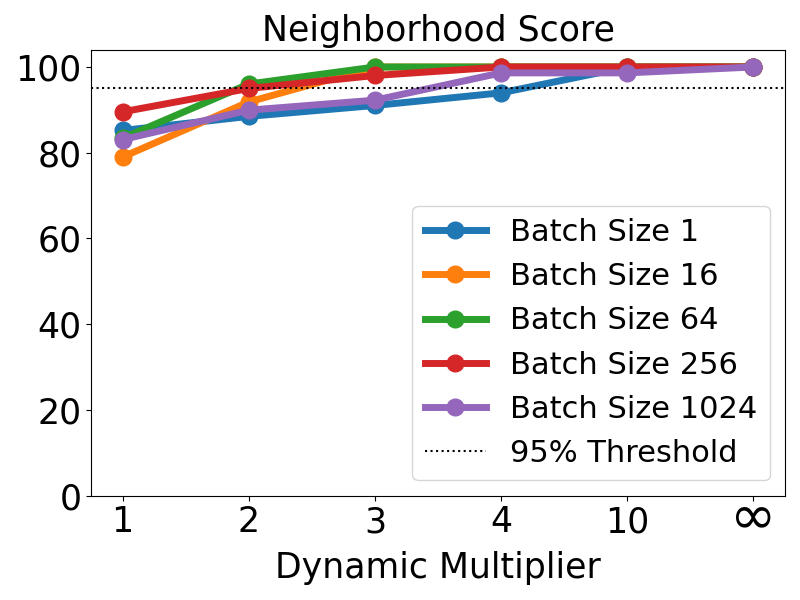}
        \caption{Neighborhood score}
        \label{fig:FastMEMIT-llama2-neighborhood}
    \end{subfigure}
    \caption{Performance of FastMEMIT in Llama 2 across different batch sizes}
    \label{fig:FastMEMIT-performance-llama2}
\end{figure*}

\subsection{Results}
The knowledge editing results for GPT-J (6B) with reduced precomputation are shown in Figures~\ref{fig:FastEMMET-performance-gptj} and \ref{fig:FastMEMIT-performance-gptj} for EMMET and MEMIT respectively. The results for the original EMMET and MEMIT algorithm with complete precomputation of 44 million tokens are represented on the x-axis by an "$\infty$" symbol. We present the results for different batch sizes from 1 to 1024 and the different evaluation metrics discussed in section \ref{sec:metrics}. The figures also contain a 95\% threshold line, which represents 95\% performance with respect to the full precomputation value. The exact numerical values for these figures are shared in Appendix \ref{sec:appendix}.

We can see that both FastEMMET and FastMEMIT achieve performance that is similar to or even better than the original algorithms with full precomputation, as shown by the overall score metric plots for both EMMET (Figure \ref{fig:FastEMMET-gptj-overall}) and MEMIT (Figure \ref{fig:FastMEMIT-gptj-overall}) for GPT-J. The results for GPT2-XL follow a very similar trend and are presented in the appendix (Figures \ref{fig:FastEMMET-performance-gpt2xl} and \ref{fig:FastMEMIT-performance-gpt2xl}). 

This is true despite using a significantly lower amount of precomputation. Starting at a dynamic multiplier of 2, the editing results are nearly identical to those of the original algorithms where computation is done over 44 million tokens. A dynamic multiplier of 2 means doing precomputation over 32k tokens for GPT-J, which is less than 0.08\% of the amount of precomputation required by the original algorithms. For GPT2-XL, a dynamic multiplier of 2 requires precomputation over 12.8k tokens, or 0.02\% of the original amount. This enables precomputation to finish within a few seconds, avoiding the large precomputation stage that precedes knowledge editing.

The results for Llama2-7B are shown in Figures \ref{fig:FastEMMET-performance-llama2} and \ref{fig:FastMEMIT-performance-llama2}. We see that the performance is within the 95\% threshold for EMMET even when $d_m = 2$, but MEMIT requires extra precomputation tokens to achieve comparable performance for smaller batch sizes. For MEMIT, we also observe that for smaller batch sizes from 1 to 10, the $C_{eff}$ is not invertible at low values of dynamics multiplier, suggesting that the cached hidden representations are highly correlated. We fix this with a minor regularization term which is added into the closed-form solution in equation \ref{eq:memit} \cite{akshat-regularization}. Note that this is needed only for batch sizes less than 10. With $d_m=10$, the editing performance for Llama2-7B is reliably close to the full precomputation performance for both algorithms. This requires approximately 0.25\% tokens when compared to the full pre-computation. 





\section{Related Work}\label{sec:related-work}
Knowledge editing methods can broadly be divided into two categories - in-context editing and parameter-modifying methods. In-context editing techniques, such as SERAC \cite{SERAC}, ICE \cite{ripple-effects}, MeLLo \cite{mquake} and GRACE \cite{grace}, allow updated knowledge to be added temporarily by providing new information in the model context. On the other hand, parameter-modifying knowledge editing do this by infusing new knowledge in the model weights. MEMIT \cite{MEMIT} and ROME \cite{rome} are two notable methods in this area that offer efficient solutions to directly edit the model parameters and are closely related to model interpretability. ROME introduced the idea of identifying key layers that store factual knowledge and then updating the corresponding weights to edit the model. MEMIT extended this approach by enabling batched editing, allowing multiple facts to be edited at once using a closed-form solution. These methods have been very popular and have seen a growing body of work in recent times that overcome various limitations at scale \cite{akshat-catastrophic}. These include methods like PMET \cite{PMET}, EMMET \cite{akshat-unified}, PRUNE \cite{ma2024perturbation}, AlphaEdit \cite{alphaedit}.


\section{Conclusion}
In this paper, we significantly reduce the upfront precomputation time required to cache hidden representation of a model before editing can begin for locate-then-edit methods like MEMIT, ROME and EMMET. We do this by first finding the theoretical minimum number of precomputation tokens required. We then empirically search for the optimal `minimum' number of precomputation tokens required to perform successful editing without compromising performance. \textbf{Our recommendation is to use 10 times the theoretical minimum of tokens, or to use a dynamic multiplier of 10}. Note that this number is less than 0.4\% of the originally used 44 million tokens. However, this number can further be reduced for specific models and editing algorithms as shown in our paper. This study allows editing for a new model to beging within a few minutes, saving many hours of precomputation time. 


\section{Limitations}
In our work, we present optimal number of tokens required for precomputation for popular knowledge editing methods.  We evaluate this in the setting of singular and batched editing. A recently popular mode of editing is sequential editing \cite{alphaedit}. We leave evaluation of optimal precomputation requirements for sequential editing to future work. Additionally, it has been shown that sequential editing also leads to loss of downstream performance \cite{akshat-catastrophic}. In this work, we do not analyze the relationship between the number of precomputation tokens and downstream performance, which we also leave for future work.



\bibliography{custom}

\appendix

\section{Appendix}
\label{sec:appendix}

\begin{table}[h!]
    \centering
    \begin{tabular}{|c|c|c|}
        \hline
        \textbf{Batch Size} & \textbf{Num Batches} & \textbf{Total Edits} \\ \hline
        1 & 1000 & 1000 \\ \hline
        16 & 10 & 160 \\ \hline
        64 & 5 & 320 \\ \hline
        256 & 5 & 1280 \\ \hline
        1024 & 3 & 3072 \\ \hline
    \end{tabular}
    \caption{Statistics for batch size and number of batches used to create the numbers for this paper.}
    \label{table:batch-size-stats}
\end{table}

\begin{figure*}[h!]
    \centering
    \begin{subfigure}{0.24\linewidth}
        \centering
        \includegraphics[width=\linewidth]{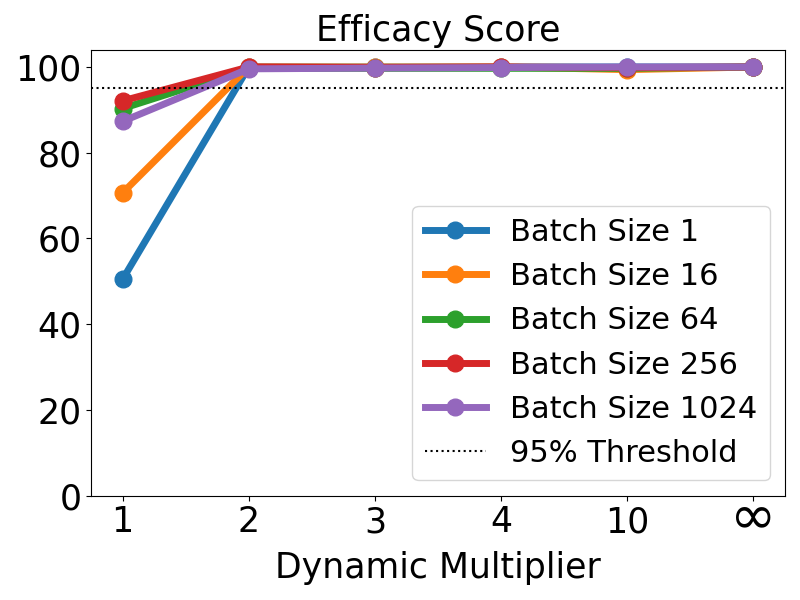}
        \caption{Efficacy score}
        \label{fig:FastEMMET-gpt2xl-efficacy}
    \end{subfigure}
    \hfill
    \begin{subfigure}{0.24\linewidth}
        \centering
        \includegraphics[width=\linewidth]{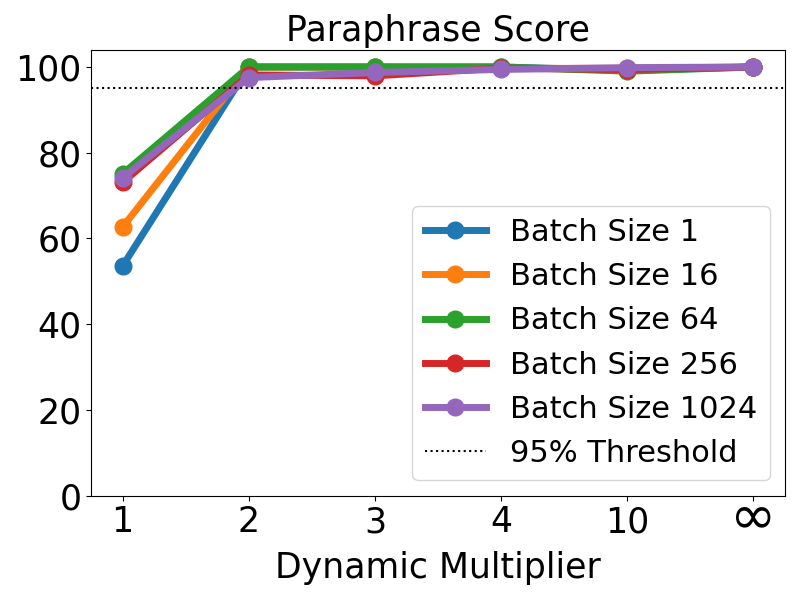}
        \caption{Paraphrase score}
        \label{fig:FastEMMET-gpt2xl-paraphrase}
    \end{subfigure}
    \hfill
    \begin{subfigure}{0.24\linewidth}
        \centering
        \includegraphics[width=\linewidth]{Images/relative_images/FastEMMET_gpt2xl_relative_ps.png}
        \caption{Neighborhood score}
        \label{fig:FastEMMET-gpt2xl-neighborhood}
    \end{subfigure}
    \hfill
    \begin{subfigure}{0.24\linewidth}
        \centering
        \includegraphics[width=\linewidth]{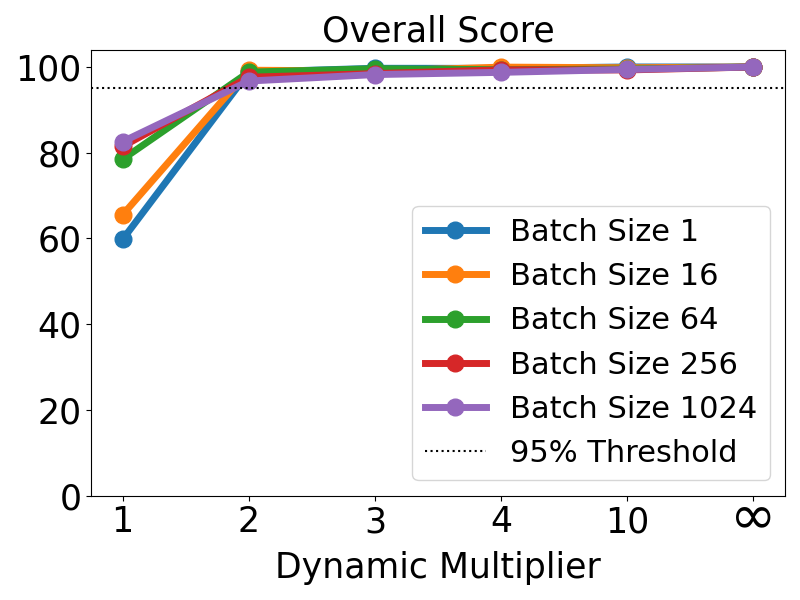}
        \caption{Overall Score}
        \label{fig:FastEMMET-gpt2xl-overall}
    \end{subfigure}
    \caption{Performance of FastEMMET in GPT2-XL across different batch sizes}
    \label{fig:FastEMMET-performance-gpt2xl}
\end{figure*}

\begin{figure*}[h!]
    \centering
    \begin{subfigure}{0.24\linewidth}
        \centering
        \includegraphics[width=\linewidth]{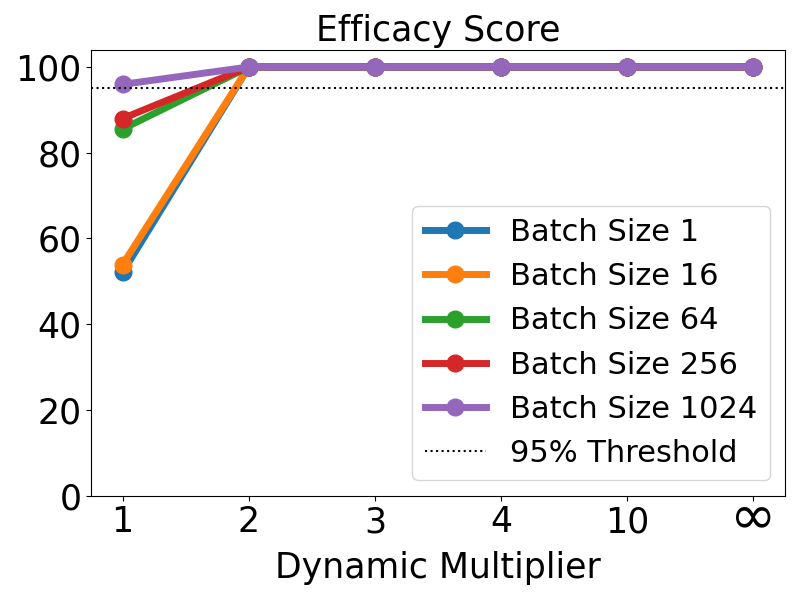}
        \caption{Efficacy score}
        \label{fig:FastMEMIT-gpt2xl-efficacy}
    \end{subfigure}
    \hfill
    \begin{subfigure}{0.24\linewidth}
        \centering
        \includegraphics[width=\linewidth]{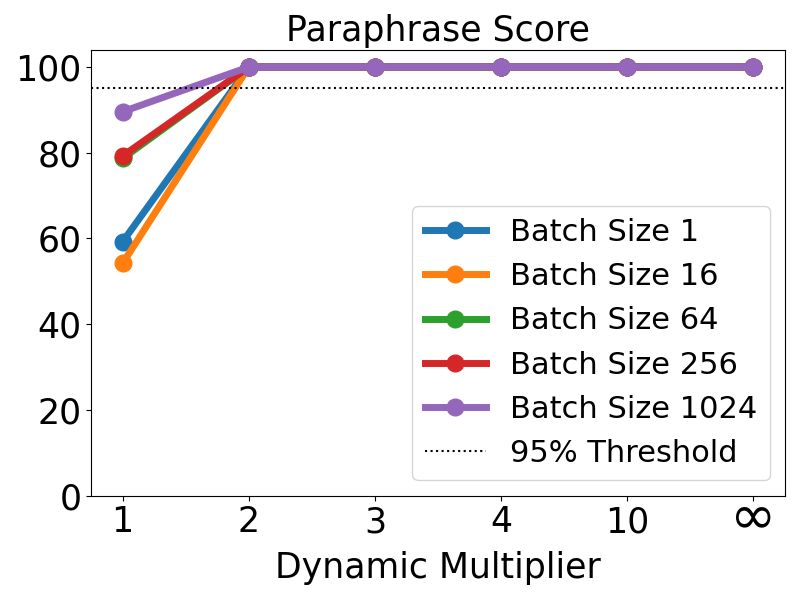}
        \caption{Paraphrase score}
        \label{fig:FastMEMIT-gpt2xl-paraphrase}
    \end{subfigure}
    \hfill
    \begin{subfigure}{0.24\linewidth}
        \centering
        \includegraphics[width=\linewidth]{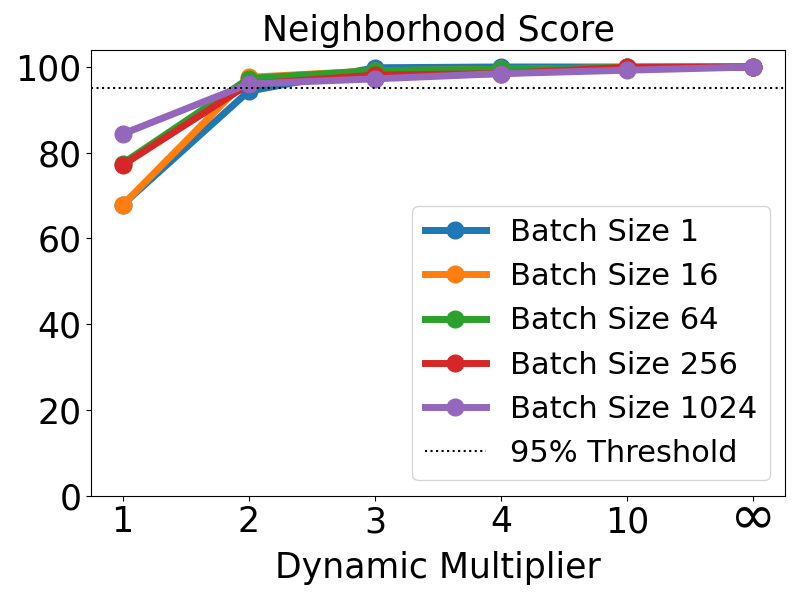}
        \caption{Neighborhood score}
        \label{fig:FastMEMIT-gpt2xl-neighborhood}
    \end{subfigure}
    \hfill
    \begin{subfigure}{0.24\linewidth}
        \centering
        \includegraphics[width=\linewidth]{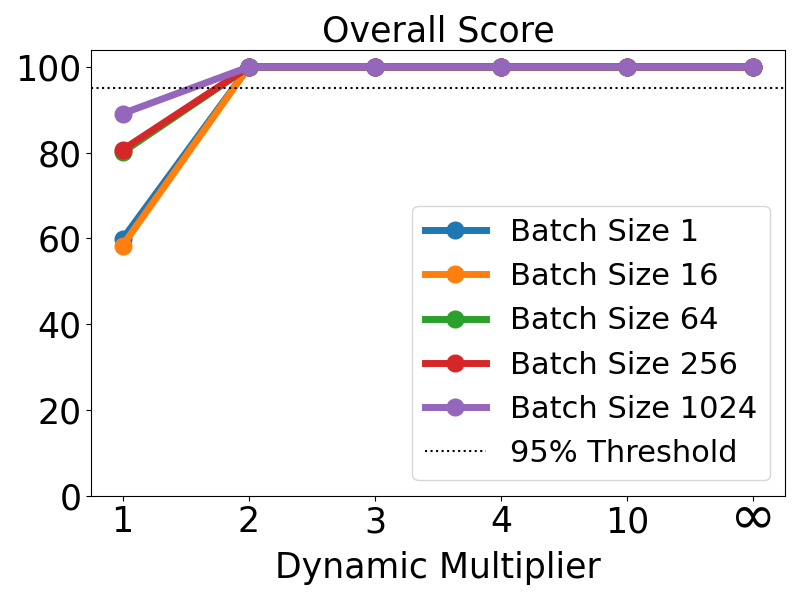}
        \caption{Overall Score}
        \label{fig:FastMEMIT-gpt2xl-overall}
    \end{subfigure}
    \caption{Performance of FastMEMIT in GPT2-XL across different batch sizes}
    \label{fig:FastMEMIT-performance-gpt2xl}
\end{figure*}
\begin{table*}
    \vskip 0.05in
    \centering
    \scriptsize
    \setlength\tabcolsep{6pt} 
    \setlength\extrarowheight{1pt}
    \begin{tabular}{c|cc|cc|cc|cc} 
        \hline
        \multirow{2}{*}{\textsc{Batch Size}} & 
        \multicolumn{2}{c|}{\textsc{ES (Efficacy)}} & 
        \multicolumn{2}{c|}{\textsc{PS (Generalization)}} & 
        \multicolumn{2}{c|}{\textsc{NS (Locality)}} & 
        \multicolumn{2}{c}{\textsc{S (Score)}} \\
        \cline{2-9}
        & \textsc{EMMET} & \textsc{FastEMMET} & \textsc{EMMET} & \textsc{FastEMMET} & \textsc{EMMET} & \textsc{FastEMMET} & \textsc{EMMET} & \textsc{FastEMMET} \\
        \hline
        1    & 99.9  & 50.4 & 94.0  & 50.35  & 65.59 & 49.09 & 83.57 & 49.93 \\ 
        16   & 100.0 & 70.62 & 96.25 & 60.31 & 74.19 & 47.69 & 88.57 & 58.01 \\
        64   & 100.0 & 90.31 & 95.94 & 72.03 & 73.56 & 54.37 & 88.18 & 69.20 \\
        256  & 99.84 & 91.95 & 96.13 & 70.23 & 67.14 & 54.83 & 84.95 & 69.20 \\
        1024 & 99.32 & 86.72 & 92.15 & 91.62 & 68.28 & 54.11 & 81.35 & 67.17 \\
        \hline
    \end{tabular}
    \caption{Comparison between EMMET and FastEMMET for multiple layers with different batch sizes, dynamic multiplier = 1 in GPT2-XL on the CounterFact dataset.}
    \label{table:memit-vs-fastmemit}
    \vskip -0.0in
\end{table*}

\begin{table*}
    \vskip 0.05in
    \centering
    \scriptsize
    \setlength\tabcolsep{6pt} 
    \setlength\extrarowheight{1pt}
    \begin{tabular}{c|cc|cc|cc|cc} 
        \hline
        \multirow{2}{*}{\textsc{Batch Size}} & 
        \multicolumn{2}{c|}{\textsc{ES (Efficacy)}} & 
        \multicolumn{2}{c|}{\textsc{PS (Generalization)}} & 
        \multicolumn{2}{c|}{\textsc{NS (Locality)}} & 
        \multicolumn{2}{c}{\textsc{S (Score)}} \\
        \cline{2-9}
        & \textsc{EMMET} & \textsc{FastEMMET} & \textsc{EMMET} & \textsc{FastEMMET} & \textsc{EMMET} & \textsc{FastEMMET} & \textsc{EMMET} & \textsc{FastEMMET} \\
        \hline
        1    & 99.9  & 100.0 & 94.0  & 94.25  & 65.59 & 63.61 & 83.57 & 82.57 \\ 
        16   & 100.0 & 100.0 & 96.25 & 96.88 & 74.19 & 72.44 & 88.57 & 87.90 \\
        64   & 100.0 & 99.69 & 95.94 & 96.88 & 73.56 & 71.03 & 88.18 & 87.12 \\
        256  & 99.84 & 99.84 & 96.13 & 94.34 & 67.14 & 64.26 & 84.95 & 82.92 \\
        1024 & 99.32 & 98.89 & 92.15 & 89.86 & 68.28 & 59.23 & 81.35 & 78.69 \\
        \hline
    \end{tabular}
    \caption{Comparison between EMMET and FastEMMET for multiple layers with different batch sizes, dynamic multiplier = 2 in GPT2-XL on the CounterFact dataset.}
    \label{table:memit-vs-fastmemit}
    \vskip -0.0in
\end{table*}

\begin{table*}
    \vskip 0.05in
    \centering
    \scriptsize
    \setlength\tabcolsep{6pt} 
    \setlength\extrarowheight{1pt}
    \begin{tabular}{c|cc|cc|cc|cc} 
        \hline
        \multirow{2}{*}{\textsc{Batch Size}} & 
        \multicolumn{2}{c|}{\textsc{ES (Efficacy)}} & 
        \multicolumn{2}{c|}{\textsc{PS (Generalization)}} & 
        \multicolumn{2}{c|}{\textsc{NS (Locality)}} & 
        \multicolumn{2}{c}{\textsc{S (Score)}} \\
        \cline{2-9}
        & \textsc{EMMET} & \textsc{FastEMMET} & \textsc{EMMET} & \textsc{FastEMMET} & \textsc{EMMET} & \textsc{FastEMMET} & \textsc{EMMET} & \textsc{FastEMMET} \\
        \hline
        1    & 99.9  & 100.0 & 94.0  & 94.35  & 65.59 & 64.91 & 83.57 & 83.32 \\ 
        16   & 100.0 & 100.0 & 96.25 & 96.25 & 74.19 & 72.44 & 88.57 & 87.73 \\
        64   & 100.0 & 99.69 & 95.94 & 97.19 & 73.56 & 72.28 & 88.18 & 87.83 \\
        256  & 99.84 & 99.69 & 96.13 & 94.14 & 67.14 & 65.72 & 84.95 & 83.63 \\
        1024 & 99.32 & 99.06 & 92.15 & 90.97 & 68.28 & 60.79 & 81.35 & 79.91 \\
        \hline
    \end{tabular}
    \caption{Comparison between EMMET and FastEMMET for multiple layers with different batch sizes, dynamic multiplier = 3 in GPT2-XL on the CounterFact dataset.}
    \label{table:memit-vs-fastmemit}
    \vskip -0.0in
\end{table*}

\begin{table*}
    \vskip 0.05in
    \centering
    \scriptsize
    \setlength\tabcolsep{6pt} 
    \setlength\extrarowheight{1pt}
    \begin{tabular}{c|cc|cc|cc|cc} 
        \hline
        \multirow{2}{*}{\textsc{Batch Size}} & 
        \multicolumn{2}{c|}{\textsc{ES (Efficacy)}} & 
        \multicolumn{2}{c|}{\textsc{PS (Generalization)}} & 
        \multicolumn{2}{c|}{\textsc{NS (Locality)}} & 
        \multicolumn{2}{c}{\textsc{S (Score)}} \\
        \cline{2-9}
        & \textsc{EMMET} & \textsc{FastEMMET} & \textsc{EMMET} & \textsc{FastEMMET} & \textsc{EMMET} & \textsc{FastEMMET} & \textsc{EMMET} & \textsc{FastEMMET} \\
        \hline
        1    & 99.9  & 100.0 & 94.0  & 94.2  & 65.59 & 64.95 & 83.57 & 83.30 \\ 
        16   & 100.0 & 100.0 & 96.25 & 95.94 & 74.19 & 74.31 & 88.57 & 88.54 \\
        64   & 100.0 & 99.69 & 95.94 & 96.56 & 73.56 & 72.10 & 88.18 & 87.61 \\
        256  & 99.84 & 99.92 & 96.13 & 95.86 & 67.14 & 66.18 & 84.95 & 84.38 \\
        1024 & 99.32 & 99.19 & 92.15 & 91.62 & 62.68 & 61.18 & 81.35 & 80.33 \\
        \hline
    \end{tabular}
    \caption{Comparison between EMMET and FastEMMET for multiple layers with different batch sizes, dynamic multiplier = 4 in GPT2-XL on the CounterFact dataset.}
    \label{table:memit-vs-fastmemit}
    \vskip -0.0in
\end{table*}

\begin{table*}
    \vskip 0.05in
    \centering
    \scriptsize
    \setlength\tabcolsep{6pt} 
    \setlength\extrarowheight{1pt}
    \begin{tabular}{c|cc|cc|cc|cc} 
        \hline
        \multirow{2}{*}{\textsc{Batch Size}} & 
        \multicolumn{2}{c|}{\textsc{ES (Efficacy)}} & 
        \multicolumn{2}{c|}{\textsc{PS (Generalization)}} & 
        \multicolumn{2}{c|}{\textsc{NS (Locality)}} & 
        \multicolumn{2}{c}{\textsc{S (Score)}} \\
        \cline{2-9}
        & \textsc{EMMET} & \textsc{FastEMMET} & \textsc{EMMET} & \textsc{FastEMMET} & \textsc{EMMET} & \textsc{FastEMMET} & \textsc{EMMET} & \textsc{FastEMMET} \\
        \hline
        1    & 99.9  & 100.0 & 94.0  & 93.25  & 65.59 & 66.55 & 83.57 & 83.89 \\ 
        16   & 100.0 & 99.38 & 96.25 & 95.94 & 74.19 & 74.38 & 88.57 & 88.41 \\
        64   & 100.0 & 99.69 & 95.94 & 95.0 & 73.56 & 73.38 & 88.18 & 87.75 \\
        256  & 99.84 & 99.69 & 96.13 & 95.55 & 67.14 & 66.4 & 84.95 & 84.37 \\
        1024 & 99.38 & 99.19 & 92.15 & 92.01 & 62.68 & 61.88 & 81.35 & 80.88 \\
        \hline
    \end{tabular}
    \caption{Comparison between EMMET and FastEMMET for multiple layers with different batch sizes, dynamic multiplier = 10 in GPT2-XL on the CounterFact dataset.}
    \label{table:memit-vs-fastmemit}
    \vskip -0.0in
\end{table*}

\begin{table*}
    \vskip 0.05in
    \centering
    \scriptsize
    \setlength\tabcolsep{6pt} 
    \setlength\extrarowheight{1pt}
    \begin{tabular}{c|cc|cc|cc|cc}
        \hline
        \multirow{2}{*}{\textsc{Batch Size}} & 
        \multicolumn{2}{c|}{\textsc{ES (Efficacy)}} & 
        \multicolumn{2}{c|}{\textsc{PS (Generalization)}} & 
        \multicolumn{2}{c|}{\textsc{NS (Locality)}} & 
        \multicolumn{2}{c}{\textsc{S (Score)}} \\
        \cline{2-9}
        & \textsc{MEMIT} & \textsc{FastMEMIT} & \textsc{MEMIT} & \textsc{FastMEMIT} & \textsc{MEMIT} & \textsc{FastMEMIT} & \textsc{MEMIT} & \textsc{FastMEMIT} \\
        \hline
        1    & 97.2  & 50.6  & 86.55 & 51.25  & 71.14 & 48.28 & 83.56 & 50.01 \\ 
        16   & 96.25 & 51.88 & 82.5  & 44.69 & 78.19 & 53.0 & 84.98 & 49.57 \\
        64   & 97.19 & 83.12 & 86.41 & 67.97 & 77.56 & 60.09 & 86.31 & 69.14 \\
        256  & 96.88 & 85.23 & 86.48 & 56.17 & 72.86 & 68.52 & 84.24 & 67.98 \\
        1024 & 95.48 & 91.6 & 84.94 & 76.07 & 70.38 & 59.37 & 82.29 & 73.33 \\
        \hline
    \end{tabular}
    \caption{Comparison between MEMIT and FastMEMIT for multiple layers with different batch sizes, dynamic multiplier = 1 in GPT2-XL on the CounterFact dataset.}
    \label{table:memit-vs-fastmemit}
    \vskip -0.0in
\end{table*}

\begin{table*}
    \vskip 0.05in
    \centering
    \scriptsize
    \setlength\tabcolsep{6pt} 
    \setlength\extrarowheight{1pt}
    \begin{tabular}{c|cc|cc|cc|cc}
        \hline
        \multirow{2}{*}{\textsc{Batch Size}} & 
        \multicolumn{2}{c|}{\textsc{ES (Efficacy)}} & 
        \multicolumn{2}{c|}{\textsc{PS (Generalization)}} & 
        \multicolumn{2}{c|}{\textsc{NS (Locality)}} & 
        \multicolumn{2}{c}{\textsc{S (Score)}} \\
        \cline{2-9}
        & \textsc{MEMIT} & \textsc{FastMEMIT} & \textsc{MEMIT} & \textsc{FastMEMIT} & \textsc{MEMIT} & \textsc{FastMEMIT} & \textsc{MEMIT} & \textsc{FastMEMIT} \\
        \hline
        1    & 97.2  & 100.0  & 86.55 & 92.45  & 71.14 & 67.12 & 83.56 & 83.99 \\ 
        16   & 96.25 & 100.0 & 82.5  & 92.19 & 78.19 & 76.31 & 84.98 & 88.36 \\
        64   & 97.19 & 100.0 & 86.41 & 95.16 & 77.56 & 75.47 & 86.31 & 88.86 \\
        256  & 96.88 & 99.77 & 86.48 & 94.3 & 72.86 & 69.94 & 84.24 & 85.89 \\
        1024 & 95.48 & 99.38 & 84.94 & 92.24 & 70.38 & 67.61 & 82.29 & 84.04 \\
        \hline
    \end{tabular}
    \caption{Comparison between MEMIT and FastMEMIT for multiple layers with different batch sizes, dynamic multiplier = 2 in GPT2-XL on the CounterFact dataset.}
    \label{table:memit-vs-fastmemit}
    \vskip -0.0in
\end{table*}

\begin{table*}
    \vskip 0.05in
    \centering
    \scriptsize
    \setlength\tabcolsep{6pt} 
    \setlength\extrarowheight{1pt}
    \begin{tabular}{c|cc|cc|cc|cc}
        \hline
        \multirow{2}{*}{\textsc{Batch Size}} & 
        \multicolumn{2}{c|}{\textsc{ES (Efficacy)}} & 
        \multicolumn{2}{c|}{\textsc{PS (Generalization)}} & 
        \multicolumn{2}{c|}{\textsc{NS (Locality)}} & 
        \multicolumn{2}{c}{\textsc{S (Score)}} \\
        \cline{2-9}
        & \textsc{MEMIT} & \textsc{FastMEMIT} & \textsc{MEMIT} & \textsc{FastMEMIT} & \textsc{MEMIT} & \textsc{FastMEMIT} & \textsc{MEMIT} & \textsc{FastMEMIT} \\
        \hline
        1    & 97.2  & 99.6  & 86.55 & 91.1  & 71.14 & 71.01 & 83.56 & 85.47 \\ 
        16   & 96.25 & 100.0 & 82.5  & 90.00 & 78.19 & 77.50 & 84.98 & 88.20 \\
        64   & 97.19 & 98.44 & 86.41 & 91.09 & 77.56 & 76.91 & 86.31 & 87.88 \\
        256  & 96.88 & 99.06 & 86.48 & 92.70 & 72.86 & 71.48 & 84.24 & 86.03 \\
        1024 & 95.48 & 98.76 & 84.94 & 90.92 & 70.38 & 68.43 & 82.29 & 83.94 \\
        \hline
    \end{tabular}
    \caption{Comparison between MEMIT and FastMEMIT for multiple layers with different batch sizes, dynamic multiplier = 3 in GPT2-XL on the CounterFact dataset.}
    \label{table:memit-vs-fastmemit}
    \vskip -0.0in
\end{table*}

\begin{table*}
    \vskip 0.05in
    \centering
    \scriptsize
    \setlength\tabcolsep{6pt} 
    \setlength\extrarowheight{1pt}
    \begin{tabular}{c|cc|cc|cc|cc}
        \hline
        \multirow{2}{*}{\textsc{Batch Size}} & 
        \multicolumn{2}{c|}{\textsc{ES (Efficacy)}} & 
        \multicolumn{2}{c|}{\textsc{PS (Generalization)}} & 
        \multicolumn{2}{c|}{\textsc{NS (Locality)}} & 
        \multicolumn{2}{c}{\textsc{S (Score)}} \\
        \cline{2-9}
        & \textsc{MEMIT} & \textsc{FastMEMIT} & \textsc{MEMIT} & \textsc{FastMEMIT} & \textsc{MEMIT} & \textsc{FastMEMIT} & \textsc{MEMIT} & \textsc{FastMEMIT} \\
        \hline
        1    & 97.2  & 99.4  & 86.55 & 91.3  & 71.14 & 71.54 & 83.56 & 85.73 \\ 
        16   & 96.25 & 98.75 & 82.5  & 86.56 & 78.19 & 77.75 & 84.98 & 86.85 \\
        64   & 97.19 & 98.44 & 86.41 & 90.16 & 77.56 & 77.12 & 86.31 & 87.67 \\
        256  & 96.88 & 99.06 & 86.48 & 90.43 & 72.86 & 71.73 & 84.24 & 85.48 \\
        1024 & 95.48 & 98.37 & 84.94 & 89.01 & 70.38 & 69.29 & 82.29 & 83.72 \\
        \hline
    \end{tabular}
    \caption{Comparison between MEMIT and FastMEMIT for multiple layers with different batch sizes, dynamic multiplier = 4 in GPT2-XL on the CounterFact dataset.}
    \label{table:memit-vs-fastmemit}
    \vskip -0.0in
\end{table*}

\begin{table*}
    \vskip 0.05in
    \centering
    \scriptsize
    \setlength\tabcolsep{6pt} 
    \setlength\extrarowheight{1pt}
    \begin{tabular}{c|cc|cc|cc|cc}
        \hline
        \multirow{2}{*}{\textsc{Batch Size}} & 
        \multicolumn{2}{c|}{\textsc{ES (Efficacy)}} & 
        \multicolumn{2}{c|}{\textsc{PS (Generalization)}} & 
        \multicolumn{2}{c|}{\textsc{NS (Locality)}} & 
        \multicolumn{2}{c}{\textsc{S (Score)}} \\
        \cline{2-9}
        & \textsc{MEMIT} & \textsc{FastMEMIT} & \textsc{MEMIT} & \textsc{FastMEMIT} & \textsc{MEMIT} & \textsc{FastMEMIT} & \textsc{MEMIT} & \textsc{FastMEMIT} \\
        \hline
        1    & 97.2  & 98.4  & 86.55 & 90.2  & 71.14 & 72.84 & 83.56 & 85.76 \\ 
        16   & 96.25 & 96.25 & 82.5  & 84.38 & 78.19 & 78.06 & 84.98 & 85.58 \\
        64   & 97.19 & 97.19 & 86.41 & 87.66 & 77.56 & 77.47 & 86.31 & 86.69 \\
        256  & 96.88 & 97.89 & 86.48 & 88.48 & 72.86 & 72.81 & 84.24 & 85.10 \\
        1024 & 95.48 & 97.04 & 84.94 & 87.65 & 70.38 & 69.85 & 82.29 & 83.26 \\
        \hline
    \end{tabular}
    \caption{Comparison between MEMIT and FastMEMIT for multiple layers with different batch sizes, dynamic multiplier = 10 in GPT2-XL on the CounterFact dataset.}
    \label{table:memit-vs-fastmemit}
    \vskip -0.0in
\end{table*}


\begin{table*}
    \vskip 0.05in
    \centering
    \scriptsize
    \setlength\tabcolsep{6pt} 
    \setlength\extrarowheight{1pt}
    \begin{tabular}{c|cc|cc|cc|cc}
        \hline
        \multirow{2}{*}{\textsc{Batch Size}} & 
        \multicolumn{2}{c|}{\textsc{ES (Efficacy)}} & 
        \multicolumn{2}{c|}{\textsc{PS (Generalization)}} & 
        \multicolumn{2}{c|}{\textsc{NS (Locality)}} & 
        \multicolumn{2}{c}{\textsc{S (Score)}} \\
        \cline{2-9}
        & \textsc{EMMET} & \textsc{FastEMMET} & \textsc{EMMET} & \textsc{FastEMMET} & \textsc{EMMET} & \textsc{FastEMMET} & \textsc{EMMET} & \textsc{FastEMMET} \\
        \hline
        1    & 99.9  & 51.6 & 94.95 & 49.85  & 77.59 & 50.01 & 89.73 & 50.47 \\ 
        16   & 100.0 & 60.0 & 93.44 & 53.44 & 81.25 & 46.69 & 90.88 & 52.81 \\
        64   & 99.69 & 58.75 & 93.91 & 52.5 & 81.78 & 52.59 & 91.16 & 54.46 \\
        256  & 99.45 & 98.12 & 94.14 & 86.6  & 78.62 & 54.66 & 89.82 & 86.6 \\
        1024 & 99.67 & 95.15 & 93.67 & 82.62 & 74.27 & 54.98 & 87.78 & 73.52 \\
        \hline
    \end{tabular}
    \caption{Comparison between EMMET and FastEMMET for multiple layers with different batch sizes, dynamic multiplier = 1 in GPT-J on the CounterFact dataset.}
    \label{table:memit-vs-fastmemit}
    \vskip -0.0in
\end{table*}

\begin{table*}
    \vskip 0.05in
    \centering
    \scriptsize
    \setlength\tabcolsep{6pt} 
    \setlength\extrarowheight{1pt}
    \begin{tabular}{c|cc|cc|cc|cc}
        \hline
        \multirow{2}{*}{\textsc{Batch Size}} & 
        \multicolumn{2}{c|}{\textsc{ES (Efficacy)}} & 
        \multicolumn{2}{c|}{\textsc{PS (Generalization)}} & 
        \multicolumn{2}{c|}{\textsc{NS (Locality)}} & 
        \multicolumn{2}{c}{\textsc{S (Score)}} \\
        \cline{2-9}
        & \textsc{EMMET} & \textsc{FastEMMET} & \textsc{EMMET} & \textsc{FastEMMET} & \textsc{EMMET} & \textsc{FastEMMET} & \textsc{EMMET} & \textsc{FastEMMET} \\
        \hline
        1    & 99.9  & 100.0 & 94.95 & 97.4  & 77.59 & 71.81 & 89.73 & 87.73 \\ 
        16   & 100.0 & 100.0 & 93.44 & 96.56 & 81.25 & 79.88 & 90.88 & 91.25 \\
        64   & 99.69 & 100.0 & 93.91 & 96.88 & 81.78 & 80.03 & 91.16 & 91.41 \\
        256  & 99.45 & 99.84 & 94.14 & 97.19  & 78.62 & 74.18 & 89.82 & 88.79 \\
        1024 & 99.67 & 99.8 & 93.67 & 96.35 & 74.27 & 68.01 & 87.78 & 85.46 \\
        \hline
    \end{tabular}
    \caption{Comparison between EMMET and FastEMMET for multiple layers with different batch sizes, dynamic multiplier = 2 in GPT-J on the CounterFact dataset.}
    \label{table:memit-vs-fastmemit}
    \vskip -0.0in
\end{table*}

\begin{table*}
    \vskip 0.05in
    \centering
    \scriptsize
    \setlength\tabcolsep{6pt} 
    \setlength\extrarowheight{1pt}
    \begin{tabular}{c|cc|cc|cc|cc}
        \hline
        \multirow{2}{*}{\textsc{Batch Size}} & 
        \multicolumn{2}{c|}{\textsc{ES (Efficacy)}} & 
        \multicolumn{2}{c|}{\textsc{PS (Generalization)}} & 
        \multicolumn{2}{c|}{\textsc{NS (Locality)}} & 
        \multicolumn{2}{c}{\textsc{S (Score)}} \\
        \cline{2-9}
        & \textsc{EMMET} & \textsc{FastEMMET} & \textsc{EMMET} & \textsc{FastEMMET} & \textsc{EMMET} & \textsc{FastEMMET} & \textsc{EMMET} & \textsc{FastEMMET} \\
        \hline
        1    & 99.9  & 100.0 & 94.95 & 96.65  & 77.59 & 75.2 & 89.73 & 89.16 \\ 
        16   & 100.0 & 100.0 & 93.44 & 94.38 & 81.25 & 80.88 & 90.88 & 91.02 \\
        64   & 99.69 & 100.0 & 93.91 & 96.56 & 81.78 & 80.72 & 91.16 & 91.61 \\
        256  & 99.45 & 99.84 & 94.14 & 96.99  & 78.62 & 75.73 & 89.82 & 89.46 \\
        1024 & 99.67 & 99.8 & 93.67 & 96.14 & 74.27 & 70.16 & 87.78 & 86.51 \\
        \hline
    \end{tabular}
    \caption{Comparison between EMMET and FastEMMET for multiple layers with different batch sizes, dynamic multiplier = 3 in GPT-J on the CounterFact dataset.}
    \label{table:memit-vs-fastmemit}
    \vskip -0.0in
\end{table*}

\begin{table*}
    \vskip 0.05in
    \centering
    \scriptsize
    \setlength\tabcolsep{6pt} 
    \setlength\extrarowheight{1pt}
    \begin{tabular}{c|cc|cc|cc|cc}
        \hline
        \multirow{2}{*}{\textsc{Batch Size}} & 
        \multicolumn{2}{c|}{\textsc{ES (Efficacy)}} & 
        \multicolumn{2}{c|}{\textsc{PS (Generalization)}} & 
        \multicolumn{2}{c|}{\textsc{NS (Locality)}} & 
        \multicolumn{2}{c}{\textsc{S (Score)}} \\
        \cline{2-9}
        & \textsc{EMMET} & \textsc{FastEMMET} & \textsc{EMMET} & \textsc{FastEMMET} & \textsc{EMMET} & \textsc{FastEMMET} & \textsc{EMMET} & \textsc{FastEMMET} \\
        \hline
        1    & 99.9  & 100.0 & 94.95 & 96.85  & 77.59 & 74.72 & 89.73 & 88.99 \\ 
        16   & 100.0 & 100.0 & 93.44 & 95.0 & 81.25 & 81.12 & 90.88 & 91.31 \\
        64   & 99.69 & 100.0 & 93.91 & 94.06 & 81.78 & 81.72 & 91.16 & 91.27 \\
        256  & 99.45 & 99.92 & 94.14 & 96.33  & 78.62 & 76.45 & 89.82 & 89.63 \\
        1024 & 99.67 & 99.74 & 93.67 & 95.8 & 74.27 & 71.03 & 87.78 & 86.84 \\
        \hline
    \end{tabular}
    \caption{Comparison between EMMET and FastEMMET for multiple layers with different batch sizes, dynamic multiplier = 4 in GPT-J on the CounterFact dataset.}
    \label{table:memit-vs-fastmemit}
    \vskip -0.0in
\end{table*}

\begin{table*}
    \vskip 0.05in
    \centering
    \scriptsize
    \setlength\tabcolsep{6pt} 
    \setlength\extrarowheight{1pt}
    \begin{tabular}{c|cc|cc|cc|cc}
        \hline
        \multirow{2}{*}{\textsc{Batch Size}} & 
        \multicolumn{2}{c|}{\textsc{ES (Efficacy)}} & 
        \multicolumn{2}{c|}{\textsc{PS (Generalization)}} & 
        \multicolumn{2}{c|}{\textsc{NS (Locality)}} & 
        \multicolumn{2}{c}{\textsc{S (Score)}} \\
        \cline{2-9}
        & \textsc{EMMET} & \textsc{FastEMMET} & \textsc{EMMET} & \textsc{FastEMMET} & \textsc{EMMET} & \textsc{FastEMMET} & \textsc{EMMET} & \textsc{FastEMMET} \\
        \hline
        1    & 99.9  & 100.0 & 94.95 & 95.7  & 77.59 & 76.55 & 89.73 & 89.51 \\ 
        16   & 100.0 & 100.0 & 93.44 & 94.06 & 81.25 & 81.19 & 90.88 & 91.05 \\
        64   & 99.69 & 100.0 & 93.91 & 94.53 & 81.78 & 81.78 & 91.16 & 91.44 \\
        256  & 99.45 & 99.77 & 94.14 & 94.8  & 78.62 & 77.27 & 89.82 & 89.51 \\
        1024 & 99.67 & 99.64 & 93.67 & 94.91 & 74.27 & 73.24 & 87.78 & 87.65 \\
        \hline
    \end{tabular}
    \caption{Comparison between EMMET and FastEMMET for multiple layers with different batch sizes, dynamic multiplier = 10 in GPT-J on the CounterFact dataset.}
    \label{table:memit-vs-fastmemit}
    \vskip -0.0in
\end{table*}


\begin{table*}
    \vskip 0.05in
    \centering
    \scriptsize
    \setlength\tabcolsep{6pt} 
    \setlength\extrarowheight{1pt}
    \begin{tabular}{c|cc|cc|cc|cc}
        \hline
        \multirow{2}{*}{\textsc{Batch Size}} & 
        \multicolumn{2}{c|}{\textsc{ES (Efficacy)}} & 
        \multicolumn{2}{c|}{\textsc{PS (Generalization)}} & 
        \multicolumn{2}{c|}{\textsc{NS (Locality)}} & 
        \multicolumn{2}{c}{\textsc{S (Score)}} \\
        \cline{2-9}
        & \textsc{MEMIT} & \textsc{FastMEMIT} & \textsc{MEMIT} & \textsc{FastMEMIT} & \textsc{MEMIT} & \textsc{FastMEMIT} & \textsc{MEMIT} & \textsc{FastMEMIT} \\
        \hline
        1    & 100.0 & 51.3 & 94.75 & 49.8 & 80.34 & 50.12 & 86.27 & 50.39 \\ 
        16   & 100.0 & 43.12 & 96.56 & 52.55 & 80.19 & 46.88 & 91.38 & 47.19 \\
        64   & 100.0 & 46.88 & 96.09 & 47.97 & 81.28 & 52.56 & 91.71 & 49.01 \\
        256  & 99.77 & 75.55 & 96.02 & 62.23 & 78.02 & 55.02 & 90.21 & 63.18 \\
        1024 & 99.74 & 76.4 & 94.66 & 65.27 & 75.65 & 56.43 & 88.73 & 65.03 \\
        \hline
    \end{tabular}
    \caption{Comparison between MEMIT and FastMEMIT for multiple layers with different batch sizes, dynamic multiplier = 1 in GPT-J on the CounterFact dataset.}
    \label{table:memit-vs-fastmemit}
    \vskip -0.0in
\end{table*}

\begin{table*}
    \vskip 0.05in
    \centering
    \scriptsize
    \setlength\tabcolsep{6pt} 
    \setlength\extrarowheight{1pt}
    \begin{tabular}{c|cc|cc|cc|cc}
        \hline
        \multirow{2}{*}{\textsc{Batch Size}} & 
        \multicolumn{2}{c|}{\textsc{ES (Efficacy)}} & 
        \multicolumn{2}{c|}{\textsc{PS (Generalization)}} & 
        \multicolumn{2}{c|}{\textsc{NS (Locality)}} & 
        \multicolumn{2}{c}{\textsc{S (Score)}} \\
        \cline{2-9}
        & \textsc{MEMIT} & \textsc{FastMEMIT} & \textsc{MEMIT} & \textsc{FastMEMIT} & \textsc{MEMIT} & \textsc{FastMEMIT} & \textsc{MEMIT} & \textsc{FastMEMIT} \\
        \hline
        1    & 100.0 & 100.0 & 94.75 & 96.9 & 80.34 & 67.79 & 86.27 & 85.53 \\ 
        16   & 100.0 & 100.0 & 96.56 & 96.88 & 80.19 & 80.38 & 91.38 & 91.56 \\
        64   & 100.0 & 100.0 & 96.09 & 96.41 & 81.28 & 80.41 & 91.71 & 91.43 \\
        256  & 99.77 & 99.77 & 96.02 & 96.88 & 78.02 & 77.16 & 90.21 & 90.07 \\
        1024 & 99.74 & 99.71 & 94.66 & 95.49 & 75.65 & 74.05 & 88.73 & 88.22 \\
        \hline
    \end{tabular}
    \caption{Comparison between MEMIT and FastMEMIT for multiple layers with different batch sizes, dynamic multiplier = 2 in GPT-J on the CounterFact dataset.}
    \label{table:memit-vs-fastmemit}
    \vskip -0.0in
\end{table*}

\begin{table*}
    \vskip 0.05in
    \centering
    \scriptsize
    \setlength\tabcolsep{6pt} 
    \setlength\extrarowheight{1pt}
    \begin{tabular}{c|cc|cc|cc|cc}
        \hline
        \multirow{2}{*}{\textsc{Batch Size}} & 
        \multicolumn{2}{c|}{\textsc{ES (Efficacy)}} & 
        \multicolumn{2}{c|}{\textsc{PS (Generalization)}} & 
        \multicolumn{2}{c|}{\textsc{NS (Locality)}} & 
        \multicolumn{2}{c}{\textsc{S (Score)}} \\
        \cline{2-9}
        & \textsc{MEMIT} & \textsc{FastMEMIT} & \textsc{MEMIT} & \textsc{FastMEMIT} & \textsc{MEMIT} & \textsc{FastMEMIT} & \textsc{MEMIT} & \textsc{FastMEMIT} \\
        \hline
        1    & 100.0 & 100.0 & 94.75 & 96.75 & 80.34 & 72.72 & 86.27 & 88.00 \\ 
        16   & 100.0 & 100.0 & 96.56 & 97.19 & 80.19 & 80.62 & 91.38 & 91.76 \\
        64   & 100.0 & 100.0 & 96.09 & 97.03 & 81.28 & 81.09 & 91.71 & 91.91 \\
        256  & 99.77 & 99.84 & 96.02 & 96.56 & 78.02 & 77.77 & 90.21 & 90.27 \\
        1024 & 99.74 & 99.71 & 94.66 & 95.48 & 75.65 & 74.88 & 88.73 & 88.60 \\
        \hline
    \end{tabular}
    \caption{Comparison between MEMIT and FastMEMIT for multiple layers with different batch sizes, dynamic multiplier = 3 in GPT-J on the CounterFact dataset.}
    \label{table:memit-vs-fastmemit}
    \vskip -0.0in
\end{table*}

\begin{table*}
    \vskip 0.05in
    \centering
    \scriptsize
    \setlength\tabcolsep{6pt} 
    \setlength\extrarowheight{1pt}
    \begin{tabular}{c|cc|cc|cc|cc}
        \hline
        \multirow{2}{*}{\textsc{Batch Size}} & 
        \multicolumn{2}{c|}{\textsc{ES (Efficacy)}} & 
        \multicolumn{2}{c|}{\textsc{PS (Generalization)}} & 
        \multicolumn{2}{c|}{\textsc{NS (Locality)}} & 
        \multicolumn{2}{c}{\textsc{S (Score)}} \\
        \cline{2-9}
        & \textsc{MEMIT} & \textsc{FastMEMIT} & \textsc{MEMIT} & \textsc{FastMEMIT} & \textsc{MEMIT} & \textsc{FastMEMIT} & \textsc{MEMIT} & \textsc{FastMEMIT} \\
        \hline
        1    & 100.0 & 100.0 & 94.75 & 96.45 & 80.34 & 73.78 & 86.27 & 88.43 \\ 
        16   & 100.0 & 100.0 & 96.56 & 96.56 & 80.19 & 80.62 & 91.38 & 91.57 \\
        64   & 100.0 & 100.0 & 96.09 & 96.88 & 81.28 & 80.53 & 91.71 & 91.63 \\
        256  & 99.77 & 99.84 & 96.02 & 96.52 & 78.02 & 78.05 & 90.21 & 90.39 \\
        1024 & 99.74 & 99.74 & 94.66 & 95.12 & 75.65 & 75.65 & 88.73 & 88.89 \\
        \hline
    \end{tabular}
    \caption{Comparison between MEMIT and FastMEMIT for multiple layers with different batch sizes, dynamic multiplier = 4 in GPT-J on the CounterFact dataset.}
    \label{table:memit-vs-fastmemit}
    \vskip -0.0in
\end{table*}

\begin{table*}
    \vskip 0.05in
    \centering
    \scriptsize
    \setlength\tabcolsep{6pt} 
    \setlength\extrarowheight{1pt}
    \begin{tabular}{c|cc|cc|cc|cc}
        \hline
        \multirow{2}{*}{\textsc{Batch Size}} & 
        \multicolumn{2}{c|}{\textsc{ES (Efficacy)}} & 
        \multicolumn{2}{c|}{\textsc{PS (Generalization)}} & 
        \multicolumn{2}{c|}{\textsc{NS (Locality)}} & 
        \multicolumn{2}{c}{\textsc{S (Score)}} \\
        \cline{2-9}
        & \textsc{MEMIT} & \textsc{FastMEMIT} & \textsc{MEMIT} & \textsc{FastMEMIT} & \textsc{MEMIT} & \textsc{FastMEMIT} & \textsc{MEMIT} & \textsc{FastMEMIT} \\
        \hline
        1    & 100.0 & 100.0 & 94.75 & 95.65 & 80.34 & 74.85 & 86.27 & 88.71 \\ 
        16   & 100.0 & 100.0 & 96.56 & 96.25 & 80.19 & 80.75 & 91.38 & 91.53 \\
        64   & 100.0 & 100.0 & 96.09 & 96.72 & 81.28 & 81.31 & 91.71 & 91.91 \\
        256  & 99.77 & 99.69 & 96.02 & 95.43 & 78.02 & 78.38 & 90.21 & 90.17 \\
        1024 & 99.74 & 99.71 & 94.66 & 94.61 & 75.65 & 76.12 & 88.73 & 88.92 \\
        \hline
    \end{tabular}
    \caption{Comparison between MEMIT and FastMEMIT for multiple layers with different batch sizes, dynamic multiplier = 10 in GPT-J on the CounterFact dataset.}
    \label{table:memit-vs-fastmemit}
    \vskip -0.0in
\end{table*}

\begin{table*}
    \vskip 0.05in
    \centering
    \scriptsize
    \setlength\tabcolsep{6pt} 
    \setlength\extrarowheight{1pt}
    \begin{tabular}{c|cc|cc|cc|cc}
        \hline
        \multirow{2}{*}{\textsc{Batch Size}} & 
        \multicolumn{2}{c|}{\textsc{ES (Efficacy)}} & 
        \multicolumn{2}{c|}{\textsc{PS (Generalization)}} & 
        \multicolumn{2}{c|}{\textsc{NS (Locality)}} & 
        \multicolumn{2}{c}{\textsc{S (Score)}} \\
        \cline{2-9}
        & \textsc{EMMET} & \textsc{FastEMMET} & \textsc{EMMET} & \textsc{FastEMMET} & \textsc{EMMET} & \textsc{FastEMMET} & \textsc{EMMET} & \textsc{FastEMMET} \\
        \hline
        1    & 99.5  & 51.2 & 98.5 & 49.45  & 59.0 & 50.15 & 80.75 & 50.26 \\ 
        16   & 99.38 & 52.5 & 95.62 & 49.69 & 82.94 & 51.56 & 92.08 & 51.22 \\
        64   & 98.44 & 49.38 & 97.19 & 47.66 & 78.0 & 52.03 & 90.17 & 49.62 \\
        256  & 99.61 & 75.31 & 97.89 & 60.86  & 62.1 & 51.25 & 82.51 & 60.94 \\
        1024 & 98.73 & 73.44 & 96.14 & 62.65 & 57.17 & 50.34 & 78.90 & 60.67 \\
        \hline
    \end{tabular}
    \caption{Comparison between EMMET and FastEMMET for multiple layers with different batch sizes, dynamic multiplier = 1 in Llama 2 on the CounterFact dataset.}
    \label{table:memit-vs-fastmemit}
    \vskip -0.0in
\end{table*}

\begin{table*}
    \vskip 0.05in
    \centering
    \scriptsize
    \setlength\tabcolsep{6pt} 
    \setlength\extrarowheight{1pt}
    \begin{tabular}{c|cc|cc|cc|cc}
        \hline
        \multirow{2}{*}{\textsc{Batch Size}} & 
        \multicolumn{2}{c|}{\textsc{ES (Efficacy)}} & 
        \multicolumn{2}{c|}{\textsc{PS (Generalization)}} & 
        \multicolumn{2}{c|}{\textsc{NS (Locality)}} & 
        \multicolumn{2}{c}{\textsc{S (Score)}} \\
        \cline{2-9}
        & \textsc{EMMET} & \textsc{FastEMMET} & \textsc{EMMET} & \textsc{FastEMMET} & \textsc{EMMET} & \textsc{FastEMMET} & \textsc{EMMET} & \textsc{FastEMMET} \\
        \hline
        1    & 99.5  & 99.8 & 98.5 & 99.0  & 59.0 & 49.54 & 80.75 & 74.42 \\ 
        16   & 99.38 & 99.38 & 95.62 & 98.12 & 82.94 & 72.25 & 92.08 & 87.98 \\
        64   & 98.44 & 100.0 & 97.19 & 97.81 & 78.0 & 66.09 & 90.17 & 84.85 \\
        256  & 99.61 & 98.98 & 97.89 & 95.74  & 62.1 & 53.37 & 82.51 & 76.36 \\
        1024 & 98.73 & 94.66 & 96.14 & 88.64 & 57.17 & 51.49 & 87.78 & 78.9 \\
        \hline
    \end{tabular}
    \caption{Comparison between EMMET and FastEMMET for multiple layers with different batch sizes, dynamic multiplier = 2 in Llama 2 on the CounterFact dataset.}
    \label{table:memit-vs-fastmemit}
    \vskip -0.0in
\end{table*}

\begin{table*}
    \vskip 0.05in
    \centering
    \scriptsize
    \setlength\tabcolsep{6pt} 
    \setlength\extrarowheight{1pt}
    \begin{tabular}{c|cc|cc|cc|cc}
        \hline
        \multirow{2}{*}{\textsc{Batch Size}} & 
        \multicolumn{2}{c|}{\textsc{ES (Efficacy)}} & 
        \multicolumn{2}{c|}{\textsc{PS (Generalization)}} & 
        \multicolumn{2}{c|}{\textsc{NS (Locality)}} & 
        \multicolumn{2}{c}{\textsc{S (Score)}} \\
        \cline{2-9}
        & \textsc{EMMET} & \textsc{FastEMMET} & \textsc{EMMET} & \textsc{FastEMMET} & \textsc{EMMET} & \textsc{FastEMMET} & \textsc{EMMET} & \textsc{FastEMMET} \\
        \hline
        1    & 99.5  & 99.9 & 98.5 & 99.0  & 59.0 & 52.04 & 80.75 & 76.28 \\ 
        16   & 99.38 & 98.12 & 95.62 & 96.25 & 82.94 & 77.19 & 92.08 & 89.45 \\
        64   & 98.44 & 99.69 & 97.19 & 98.44 & 78.0 & 70.94 & 90.17 & 87.49 \\
        256  & 99.61 & 99.3 & 97.89 & 97.62  & 62.1 & 56.05 & 82.51 & 78.62 \\
        1024 & 98.73 & 97.59 & 96.14 & 92.74 & 57.17 & 52.24 & 78.90 & 74.67 \\
        \hline
    \end{tabular}
    \caption{Comparison between EMMET and FastEMMET for multiple layers with different batch sizes, dynamic multiplier = 3 in Llama 2 on the CounterFact dataset.}
    \label{table:memit-vs-fastmemit}
    \vskip -0.0in
\end{table*}

\begin{table*}
    \vskip 0.05in
    \centering
    \scriptsize
    \setlength\tabcolsep{6pt} 
    \setlength\extrarowheight{1pt}
    \begin{tabular}{c|cc|cc|cc|cc}
        \hline
        \multirow{2}{*}{\textsc{Batch Size}} & 
        \multicolumn{2}{c|}{\textsc{ES (Efficacy)}} & 
        \multicolumn{2}{c|}{\textsc{PS (Generalization)}} & 
        \multicolumn{2}{c|}{\textsc{NS (Locality)}} & 
        \multicolumn{2}{c}{\textsc{S (Score)}} \\
        \cline{2-9}
        & \textsc{EMMET} & \textsc{FastEMMET} & \textsc{EMMET} & \textsc{FastEMMET} & \textsc{EMMET} & \textsc{FastEMMET} & \textsc{EMMET} & \textsc{FastEMMET} \\
        \hline
        1    & 99.5  & 99.7 & 98.5 & 98.0  & 59.0 & 54.39 & 80.75 & 77.68 \\ 
        16   & 99.38 & 100.0 & 95.62 & 97.81 & 82.94 & 79.25 & 92.08 & 91.34 \\
        64   & 98.44 & 99.69 & 97.19 & 97.81 & 78.0 & 71.53 & 90.17 & 87.62 \\
        256  & 99.61 & 99.22 & 97.89 & 96.72  & 62.1 & 57.05 & 82.51 & 79.05 \\
        1024 & 98.73 & 97.85 & 96.14 & 94.32 & 57.17 & 52.86 & 78.90 & 75.49 \\
        \hline
    \end{tabular}
    \caption{Comparison between EMMET and FastEMMET for multiple layers with different batch sizes, dynamic multiplier = 4 in Llama 2 on the CounterFact dataset.}
    \label{table:memit-vs-fastmemit}
    \vskip -0.0in
\end{table*}

\begin{table*}
    \vskip 0.05in
    \centering
    \scriptsize
    \setlength\tabcolsep{6pt} 
    \setlength\extrarowheight{1pt}
    \begin{tabular}{c|cc|cc|cc|cc}
        \hline
        \multirow{2}{*}{\textsc{Batch Size}} & 
        \multicolumn{2}{c|}{\textsc{ES (Efficacy)}} & 
        \multicolumn{2}{c|}{\textsc{PS (Generalization)}} & 
        \multicolumn{2}{c|}{\textsc{NS (Locality)}} & 
        \multicolumn{2}{c}{\textsc{S (Score)}} \\
        \cline{2-9}
        & \textsc{EMMET} & \textsc{FastEMMET} & \textsc{EMMET} & \textsc{FastEMMET} & \textsc{EMMET} & \textsc{FastEMMET} & \textsc{EMMET} & \textsc{FastEMMET} \\
        \hline
        1    & 99.5  & 99.8 & 98.5 & 98.25  & 59.0 & 56.72 & 80.75 & 79.30 \\ 
        16   & 99.38 & 99.38 & 95.62 & 96.88 & 82.94 & 81.81 & 92.08 & 92.00 \\
        64   & 98.44 & 99.69 & 97.19 & 97.5 & 78.0 & 75.53 & 90.17 & 89.47 \\
        256  & 99.61 & 99.22 & 97.89 & 98.2  & 62.1 & 59.17 & 82.51 & 80.72 \\
        1024 & 98.73 & 98.37 & 96.14 & 95.88 & 57.17 & 54.74 & 78.90 & 77.19 \\
        \hline
    \end{tabular}
    \caption{Comparison between EMMET and FastEMMET for multiple layers with different batch sizes, dynamic multiplier = 10 in Llama 2 on the CounterFact dataset.}
    \label{table:memit-vs-fastmemit}
    \vskip -0.0in
\end{table*}

\begin{table*}
    \vskip 0.05in
    \centering
    \scriptsize
    \setlength\tabcolsep{6pt} 
    \setlength\extrarowheight{1pt}
    \begin{tabular}{c|cc|cc|cc|cc}
        \hline
        \multirow{2}{*}{\textsc{Batch Size}} & 
        \multicolumn{2}{c|}{\textsc{ES (Efficacy)}} & 
        \multicolumn{2}{c|}{\textsc{PS (Generalization)}} & 
        \multicolumn{2}{c|}{\textsc{NS (Locality)}} & 
        \multicolumn{2}{c}{\textsc{S (Score)}} \\
        \cline{2-9}
        & \textsc{MEMIT} & \textsc{FastMEMIT} & \textsc{MEMIT} & \textsc{FastMEMIT} & \textsc{MEMIT} & \textsc{FastMEMIT} & \textsc{MEMIT} & \textsc{FastMEMIT} \\
        \hline
        1    & 96.6 & 53.9 & 89.4 & 54.45 & 60.82 & 51.81 & 78.98 & 53.36 \\ 
        16   & 99.38 & 46.25 & 99.38 & 44.38 & 65.12 & 51.44 & 84.55 & 47.17 \\
        64   & 98.12 & 49.69 & 97.03 & 48.59 & 61.09 & 50.94 & 81.37 & 49.72 \\
        256  & 96.33 & 79.06 & 93.59 & 65.39 & 56.85 & 50.9 & 77.60 & 63.04 \\
        1024 & 93.95 & 65.89 & 90.45 & 55.37 & 60.17 & 49.98 & 78.28 & 56.34 \\
        \hline
    \end{tabular}
    \caption{Comparison between MEMIT and FastMEMIT for multiple layers with different batch sizes, dynamic multiplier = 1 in Llama 2 on the CounterFact dataset.}
    \label{table:memit-vs-fastmemit}
    \vskip -0.0in
\end{table*}

\begin{table*}
    \vskip 0.05in
    \centering
    \scriptsize
    \setlength\tabcolsep{6pt} 
    \setlength\extrarowheight{1pt}
    \begin{tabular}{c|cc|cc|cc|cc}
        \hline
        \multirow{2}{*}{\textsc{Batch Size}} & 
        \multicolumn{2}{c|}{\textsc{ES (Efficacy)}} & 
        \multicolumn{2}{c|}{\textsc{PS (Generalization)}} & 
        \multicolumn{2}{c|}{\textsc{NS (Locality)}} & 
        \multicolumn{2}{c}{\textsc{S (Score)}} \\
        \cline{2-9}
        & \textsc{MEMIT} & \textsc{FastMEMIT} & \textsc{MEMIT} & \textsc{FastMEMIT} & \textsc{MEMIT} & \textsc{FastMEMIT} & \textsc{MEMIT} & \textsc{FastMEMIT} \\
        \hline
        1    & 96.6 & 72.5 & 89.4 & 72.5 & 60.82 & 53.81 & 78.98 & 64.97 \\ 
        16   & 99.38 & 97.5 & 99.38 & 97.5 & 65.12 & 59.81 & 84.55 & 80.57 \\
        64   & 98.12 & 95.31 & 97.03 & 95.0 & 61.09 & 58.66 & 81.37 & 78.81 \\
        256  & 96.33 & 89.14 & 93.59 & 81.68 & 56.85 & 54.0 & 77.60 & 71.46 \\
        1024 & 93.95 & 88.57 & 90.45 & 78.04 & 60.17 & 54.11 & 78.28 & 70.44 \\
        \hline
    \end{tabular}
    \caption{Comparison between MEMIT and FastMEMIT for multiple layers with different batch sizes, dynamic multiplier = 2 in Llama 2 on the CounterFact dataset.}
    \label{table:memit-vs-fastmemit}
    \vskip -0.0in
\end{table*}

\begin{table*}
    \vskip 0.05in
    \centering
    \scriptsize
    \setlength\tabcolsep{6pt} 
    \setlength\extrarowheight{1pt}
    \begin{tabular}{c|cc|cc|cc|cc}
        \hline
        \multirow{2}{*}{\textsc{Batch Size}} & 
        \multicolumn{2}{c|}{\textsc{ES (Efficacy)}} & 
        \multicolumn{2}{c|}{\textsc{PS (Generalization)}} & 
        \multicolumn{2}{c|}{\textsc{NS (Locality)}} & 
        \multicolumn{2}{c}{\textsc{S (Score)}} \\
        \cline{2-9}
        & \textsc{MEMIT} & \textsc{FastMEMIT} & \textsc{MEMIT} & \textsc{FastMEMIT} & \textsc{MEMIT} & \textsc{FastMEMIT} & \textsc{MEMIT} & \textsc{FastMEMIT} \\
        \hline
        1    & 96.6 & 90.0 & 89.4 & 85.95 & 60.82 & 55.36 & 78.98 & 73.51 \\ 
        16   & 99.38 & 99.38 & 99.38 & 97.5 & 65.12 & 67.56 & 84.55 & 85.42 \\
        64   & 98.12 & 98.12 & 97.03 & 96.56 & 61.09 & 62.13 & 81.37 & 81.87 \\
        256  & 96.33 & 95.08 & 93.59 & 91.05 & 56.85 & 55.72 & 77.60 & 76.05 \\
        1024 & 93.95 & 90.69 & 90.45 & 82.75 & 60.17 & 55.51 & 78.28 & 72.94 \\
        \hline
    \end{tabular}
    \caption{Comparison between MEMIT and FastMEMIT for multiple layers with different batch sizes, dynamic multiplier = 3 in Llama 2 on the CounterFact dataset.}
    \label{table:memit-vs-fastmemit}
    \vskip -0.0in
\end{table*}

\begin{table*}
    \vskip 0.05in
    \centering
    \scriptsize
    \setlength\tabcolsep{6pt} 
    \setlength\extrarowheight{1pt}
    \begin{tabular}{c|cc|cc|cc|cc}
        \hline
        \multirow{2}{*}{\textsc{Batch Size}} & 
        \multicolumn{2}{c|}{\textsc{ES (Efficacy)}} & 
        \multicolumn{2}{c|}{\textsc{PS (Generalization)}} & 
        \multicolumn{2}{c|}{\textsc{NS (Locality)}} & 
        \multicolumn{2}{c}{\textsc{S (Score)}} \\
        \cline{2-9}
        & \textsc{MEMIT} & \textsc{FastMEMIT} & \textsc{MEMIT} & \textsc{FastMEMIT} & \textsc{MEMIT} & \textsc{FastMEMIT} & \textsc{MEMIT} & \textsc{FastMEMIT} \\
        \hline
        1    & 96.6 & 94.9 & 89.4 & 91.05 & 60.82 & 57.15 & 78.98 & 76.88 \\ 
        16   & 99.38 & 98.75 & 99.38 & 98.44 & 65.12 & 70.19 & 84.55 & 86.87 \\
        64   & 98.12 & 98.12 & 97.03 & 96.09 & 61.09 & 63.06 & 81.37 & 82.29 \\
        256  & 96.33 & 95.08 & 93.59 & 90.51 & 56.85 & 57.33 & 77.60 & 76.90 \\
        1024 & 93.95 & 93.42 & 90.45 & 90.22 & 60.17 & 59.34 & 78.28 & 77.63 \\
        \hline
    \end{tabular}
    \caption{Comparison between MEMIT and FastMEMIT for multiple layers with different batch sizes, dynamic multiplier = 4 in Llama 2 on the CounterFact dataset.}
    \label{table:memit-vs-fastmemit}
    \vskip -0.0in
\end{table*}

\begin{table*}
    \vskip 0.05in
    \centering
    \scriptsize
    \setlength\tabcolsep{6pt} 
    \setlength\extrarowheight{1pt}
    \begin{tabular}{c|cc|cc|cc|cc}
        \hline
        \multirow{2}{*}{\textsc{Batch Size}} & 
        \multicolumn{2}{c|}{\textsc{ES (Efficacy)}} & 
        \multicolumn{2}{c|}{\textsc{PS (Generalization)}} & 
        \multicolumn{2}{c|}{\textsc{NS (Locality)}} & 
        \multicolumn{2}{c}{\textsc{S (Score)}} \\
        \cline{2-9}
        & \textsc{MEMIT} & \textsc{FastMEMIT} & \textsc{MEMIT} & \textsc{FastMEMIT} & \textsc{MEMIT} & \textsc{FastMEMIT} & \textsc{MEMIT} & \textsc{FastMEMIT} \\
        \hline
        1    & 96.6 & 98.7 & 89.4 & 93.05 & 60.82 & 67.07 & 78.98 & 83.82 \\ 
        16   & 99.38 & 98.75 & 99.38 & 98.75 & 65.12 & 75.37 & 84.55 & 89.49 \\
        64   & 98.12 & 98.44 & 97.03 & 97.5 & 61.09 & 69.88 & 81.37 & 86.39 \\
        256  & 96.33 & 97.42 & 93.59 & 94.26 & 56.85 & 57.93 & 77.60 & 78.66 \\
        1024 & 93.95 & 93.42 & 90.45 & 90.22 & 60.17 & 59.34 & 78.28 & 77.63 \\
        \hline
    \end{tabular}
    \caption{Comparison between MEMIT and FastMEMIT for multiple layers with different batch sizes, dynamic multiplier = 10 in Llama 2 on the CounterFact dataset.}
    \label{table:memit-vs-fastmemit}
    \vskip -0.0in
\end{table*}

\end{document}